\documentclass{article}

\usepackage{arxiv}
\usepackage{natbib}

\usepackage{hyperref}
\usepackage{cleveref}
\usepackage{booktabs} 
\usepackage[english]{babel}
\usepackage[utf8]{inputenc}
\usepackage{algorithm}
\usepackage[noend]{algpseudocode}
\usepackage{multirow}
\usepackage{soul,color,colortbl}
\usepackage{xcolor}
\usepackage{upgreek}
\usepackage{url}
\usepackage{verbatim}
\usepackage{xcolor}
\usepackage{booktabs}
\usepackage{upgreek}
\usepackage{mathrsfs}
\usepackage{enumerate}
\usepackage{framed}
\usepackage[framemethod=tikz]{mdframed}
\usepackage{graphicx}
\usepackage{multicol}
\usepackage{graphics}
\usepackage{cuted}
\usepackage{algorithm}
\usepackage[noend]{algpseudocode}
\usepackage{amssymb}
\usepackage{pbox}
\usepackage{amsmath}
\usepackage{pbox}
\usepackage{balance}
\usepackage{flushend}
\usepackage{amsthm}
\usepackage{stmaryrd}
\usepackage{tikz}
\usepackage{tikz-qtree}
\usepackage[justification=centering]{caption}
\usepackage{caption,booktabs}
\usepackage{subcaption}
\usepackage{graphicx}
\usepackage{tablefootnote}
\usepackage{threeparttable}

\begingroup
\catcode`[=\active \catcode`]=\active

\gdef[{\llbracket} \gdef]{\rrbracket}

\def\getdelim#1#2#3#4\relax{"#4}
\global\delcode`[=\expandafter\getdelim\llbracket\relax
\global\delcode`]=\expandafter\getdelim\rrbracket\relax
\endgroup

\title{Deep Learning for Anomaly Detection: A Survey}

\author{
  Raghavendra Chalapathy \\
  University of Sydney,\\
  Capital Markets Co-operative Research Centre (CMCRC)\\
  \texttt{rcha9612@uni.sydney.edu.au} \\
  \And
 Sanjay Chawla \\
  Qatar Computing Research Institute (QCRI), HBKU\\
  \texttt{schawla@qf.org.qa} \\
  }

\begin{document}

\maketitle

\begin{abstract}
Anomaly detection is an important problem that has been well-studied within diverse research areas and application domains. The aim of this survey is two-fold, firstly we present a structured and comprehensive overview of research methods in deep learning-based anomaly detection. Furthermore, we review the adoption of these methods for anomaly across various application domains and assess their effectiveness. We have grouped state-of-the-art deep anomaly detection research techniques into different categories based on the underlying assumptions and approach adopted. Within each category, we outline the basic anomaly detection technique, along with its variants and present key assumptions, to differentiate between normal and anomalous behavior. Besides, for each category, we also present the advantages and limitations and discuss the computational complexity of the techniques in real application domains. Finally, we outline open issues in research and challenges faced while adopting deep anomaly detection techniques for real-world problems.

\end{abstract}

\section{Introduction}

\begin{figure}[h]
\centering
\includegraphics[scale=0.5]{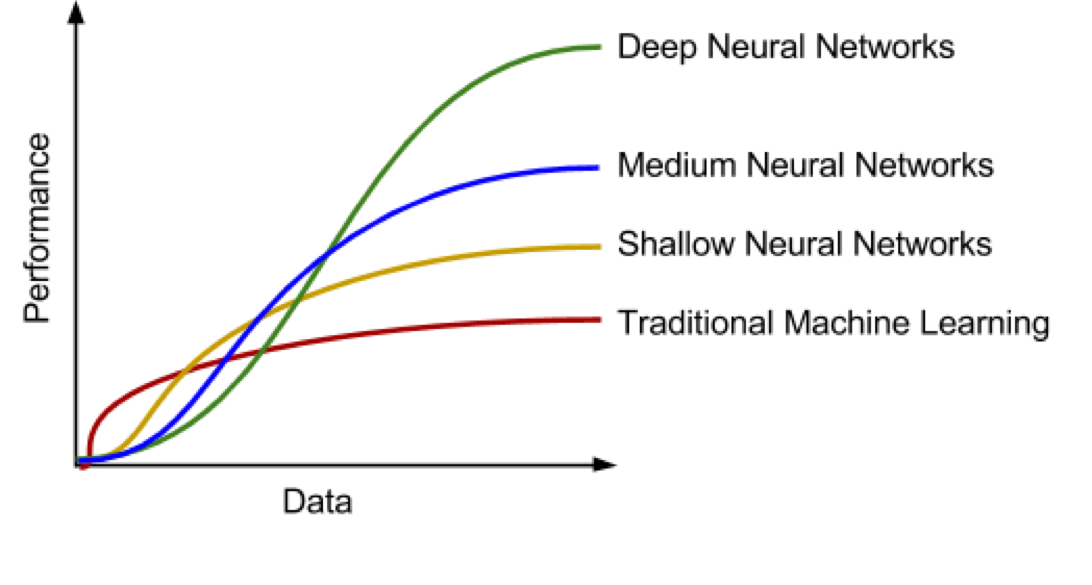}
\captionsetup{justification=centering}
\caption{Performance Comparison of Deep learning-based algorithms Vs Traditional Algorithms~\cite{deeplearningVstraditionalAlgorithms}.}
\label{fig:performanceCompare}
\end{figure}

\begin{figure}[h]
\centering
\includegraphics[scale=0.5]{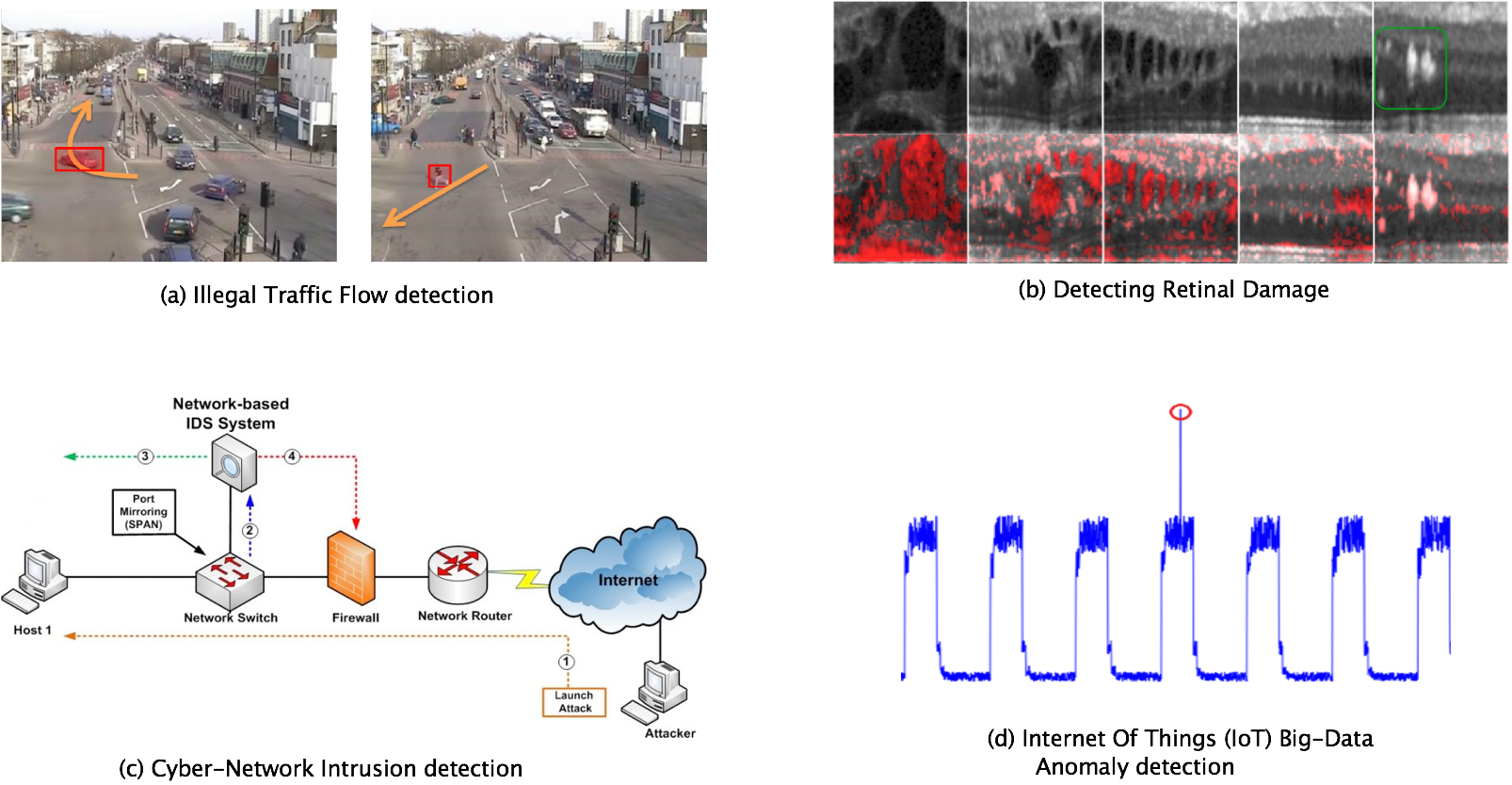}
\captionsetup{justification=centering}
\caption{Applications Deep learning-based anomaly detection algorithms.\\
(a) Video Surveillance, Image Analysis: Illegal Traffic detection~\cite{xie2017real},  (b) Health-care: Detecting Retinal Damage~\cite{schlegl2017unsupervised}\\
(c) Networks: Cyber-intrusion detection~\cite{javaid2016deep}  (d) Sensor Networks: Internet of Things (IoT) big-data anomaly detection~\cite{mohammadi2017deep} }
\label{fig:applications}
\end{figure}

A common need when analyzing real-world data-sets is determining which instances stand out as being dissimilar to all others. Such instances are known as \emph{anomalies}, and the goal of \emph{anomaly detection} (also known as \emph{outlier detection}) is to determine all such instances in a data-driven fashion~(\cite{chandola2007outlier}). Anomalies can be caused by errors in the data but sometimes are indicative of a new, previously unknown, underlying process; ~\cite{hawkins} defines an outlier as an observation that {\it deviates so significantly from other observations as to arouse suspicion that it was generated by a different mechanism.} In the broader field of machine learning, the recent years have witnessed a proliferation of deep neural networks, with unprecedented results across various application domains. Deep learning is a subset of machine learning that achieves good performance and flexibility by learning to represent the data as a nested hierarchy of concepts within layers of the neural network. Deep learning outperforms the traditional machine learning as the scale of data increases as illustrated in Figure~\ref{fig:performanceCompare}. In recent years, deep learning-based anomaly detection algorithms have become increasingly popular and have been applied for a diverse set of tasks as illustrated in Figure~\ref{fig:applications}; studies have shown that deep learning completely surpasses traditional methods ~(\cite{javaid2016deep,peng2015multi}). The aim of this survey is two-fold, firstly we present a structured and comprehensive review of research methods in deep anomaly detection (DAD). Furthermore, we also discuss the adoption of DAD methods across various application domains and assess their effectiveness.


\section{ What are anomalies?}
\begin{figure}
  \centering
  \begin{minipage}{.48\linewidth}
    \centering
      {\includegraphics[scale=0.35]{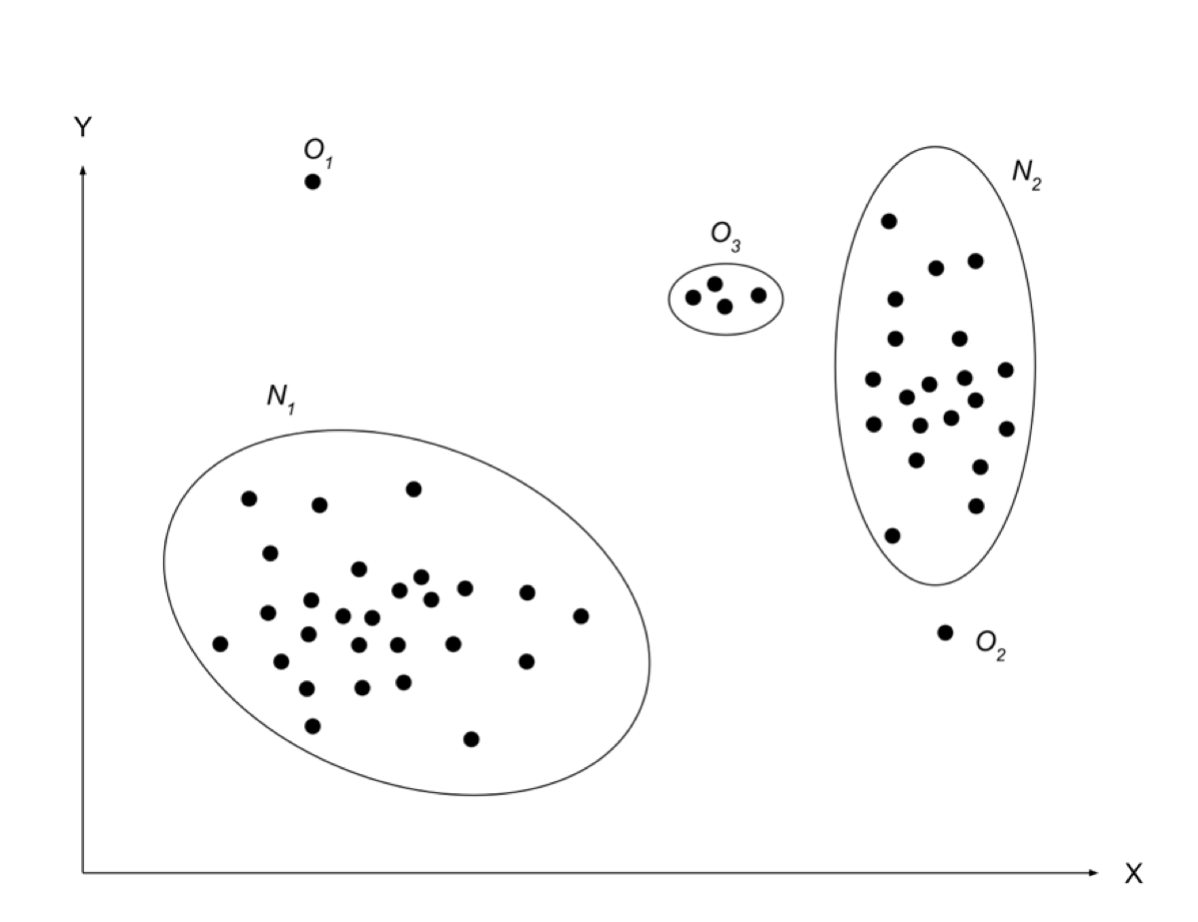}}
    \caption{Illustration of anomalies in two-dimensional data set.}
    \label{fig:anomalies}
  \end{minipage}\quad
  \begin{minipage}{.48\linewidth}
    \centering
      {\includegraphics[scale=0.35]{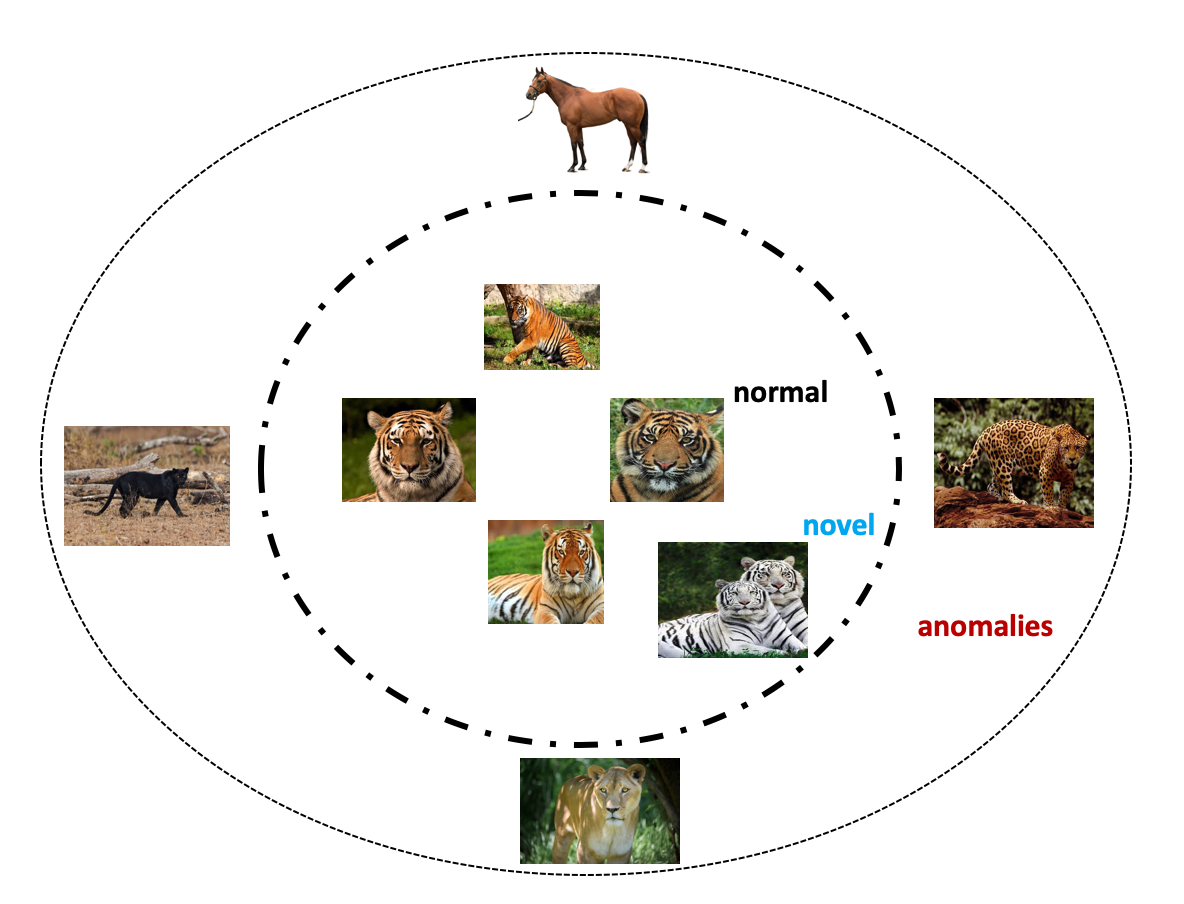}}
    \caption{Illustration of novelty in the image data set.}
    \label{fig:novelties}
  \end{minipage}
  \bigskip

\end{figure}

Anomalies  are also referred to as abnormalities, deviants, or outliers in the data mining and statistics literature~(\cite{aggarwal2013introduction}). As illustrated in Figure ~\ref{fig:anomalies}, $N_{1}$ and $N_{2}$ are regions consisting of a majority of observations and hence considered as normal data instance regions, whereas the region $O_{3}$, and data points  $O_{1}$ and $O_{2}$  are few data points which are located further away from the bulk of data points and hence are considered anomalies. arise due to several reasons, such as malicious actions, system failures, intentional fraud. These anomalies reveal exciting insights about the data and are often convey valuable information about data. Therefore, anomaly detection considered an essential step in various decision-making systems.


\section{What are novelties?}
Novelty detection is the identification of a novel (new) or unobserved patterns in the data~(\cite{miljkovic2010review}). The novelties detected are not considered as anomalous data points; instead, they are been applied to the regular data model. A novelty score may be assigned for these previously unseen data points, using a decision threshold score ~(\cite{pimentel2014review}). The points which significantly deviate from this decision threshold may be considered as anomalies or outliers. For instance, in Figure ~\ref{fig:novelties}  the images of  \textit{(white tigers)} among regular tigers may be considered as a novelty, while the image of  \textit{(horse, panther, lion, and cheetah)} are considered as anomalies. The techniques used for anomaly detection are often used for novelty detection and vice versa.

\section{Motivation and Challenges: Deep anomaly detection (DAD) techniques}
\begin{itemize}
\item Performance of traditional algorithms in detecting outliers is sub-optimal on the image (e.g. medical images) and sequence datasets since it fails to capture complex structures in the data.
\item  Need for large-scale anomaly detection: As the volume of data increases let's say to gigabytes then, it becomes nearly impossible for the traditional methods to scale to such large scale data to find outliers.
\item  Deep anomaly detection (DAD) techniques learn hierarchical discriminative features from data. This automatic feature learning capability eliminates the need of developing manual features by domain experts, therefore advocates to solve the problem end-to-end taking raw input data in domains such as text and speech recognition.
\item The boundary between normal and anomalous (erroneous) behavior is often not precisely defined  in several data domains and is continually evolving. This lack of well-defined representative normal boundary poses challenges for both conventional and deep learning-based algorithms.
\end{itemize}

\begin{table} [ht!]
\centering
\captionsetup{justification=centering}
\caption{Comparison of our Survey to Other Related Survey Articles. \\1 \textemdash Our Survey,
2 \textemdash Kwon and Donghwoon ~\cite{kwon2017survey}, 5 \textemdash John and Derek ~\cite{ball2017comprehensive}\\
3 \textemdash Kiran  and Thomas ~\cite{kiran2018overview},            6 \textemdash Mohammadi and Al-Fuqaha ~\cite{mohammadi2017deep}\\
4 \textemdash Adewumi and Andronicus ~\cite{adewumi2017survey}       7 \textemdash Geert and  Kooi et.al ~\cite{litjens2017survey}.
}
\label{tbl:surveysummary}
\scalebox{0.75}{
\begin{tabular}{ |c|c|c|c|c|c|c|c|c|c| }
\hline
 & & 1&2&3&4&5&6&7 \\
\hline
\multirow{4}{6em}{Methods  }
&Supervised &\checkmark  & & & & & & \\
&Unsupervised &\checkmark & & & & & &  \\
&Hybrid Models & \checkmark& & & & & &  \\
&one-Class Neural Networks &\checkmark & & & & & &  \\
\hline
\multirow{8}{8em}{Applications  }
&Fraud Detection&\checkmark  & & &\checkmark & & & \\
&Cyber-Intrusion Detection&\checkmark  &\checkmark & & & & & \\
&Medical Anomaly Detection&\checkmark  & & & & & &\checkmark \\
&Sensor Networks Anomaly Detection&\checkmark  & & & &\checkmark & & \\
&Internet Of Things (IoT)
 Big-data Anomaly Detection&\checkmark  & & & & & \checkmark& \\
&Log-Anomaly Detection&\checkmark  & & & & & & \\
&Video Surveillance&\checkmark & &\checkmark  & & & & \\
&Industrial Damage Detection&\checkmark & & & & & & \\
\hline
\end{tabular}}
\end{table}

\section{Related Work}
Despite the substantial advances made by deep learning methods in many machine learning problems, there
is a relative scarcity of deep learning approaches for anomaly detection. ~\cite{adewumi2017survey} provide a comprehensive survey of deep learning-based methods for fraud detection. A broad review of deep anomaly detection (DAD) techniques for cyber-intrusion detection is presented by ~\cite{kwon2017survey}. An extensive review of using DAD techniques in the medical domain is presented by  ~\cite{litjens2017survey}. An  overview of DAD techniques for the Internet of Things (IoT) and  big-data anomaly detection is introduced by ~\cite{mohammadi2017deep}. Sensor networks anomaly detection has been reviewed  by  ~\cite{ball2017comprehensive}. The state-of-the-art deep learning based methods for video anomaly detection along with various categories have been presented in~\cite{kiran2018overview}. Although there are some reviews in applying DAD techniques, there is a shortage of comparative analysis of deep learning architecture adopted for outlier detection. For instance, a substantial amount of research on anomaly detection is conducted using deep autoencoders, but there is a lack of comprehensive survey of various deep architecture's best suited for a given data-set and application domain. We hope that this survey bridges this gap and provides a comprehensive reference for researchers and engineers aspiring to leverage deep learning for anomaly detection. Table~\ref{tbl:surveysummary} shows the set of research methods and application domains covered by our survey.


\begin{figure}[h]
\centering
\includegraphics[scale=0.45]{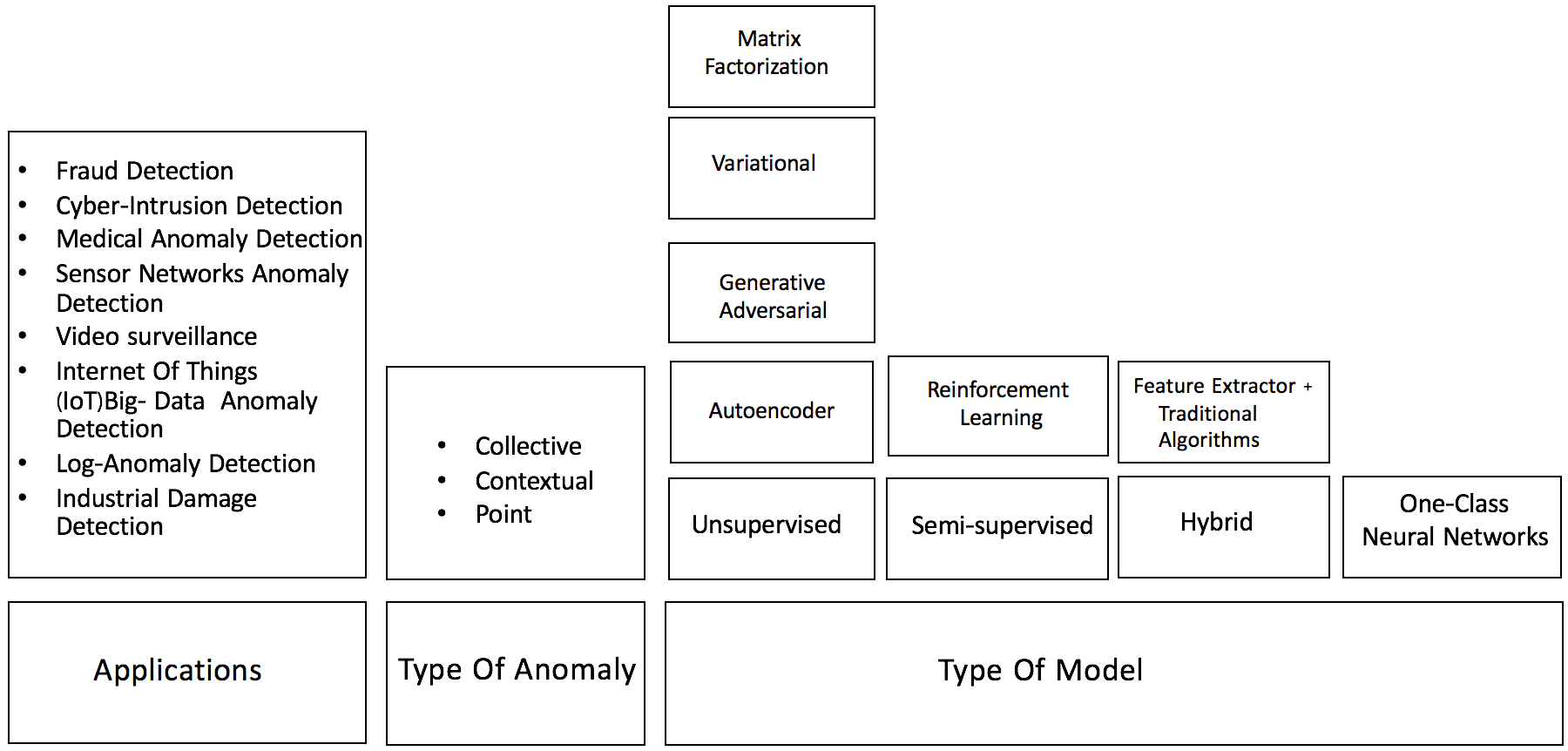}
\caption{Key components associated with deep learning-based anomaly detection technique.}
\label{fig:surveyTaxonomy}
\end{figure}

\section{ Our Contributions}
We follow the survey approach of ~(\cite{chandola2007outlier}) for deep  anomaly detection (DAD). Our survey presents a detailed and structured overview of research and applications of DAD techniques. We summarize our main contributions as follows:
\begin{itemize}
\item Most of the existing surveys on DAD techniques either focus on a particular
application domain or specific research area of interest~(\cite{kiran2018overview,mohammadi2017deep,litjens2017survey,kwon2017survey,adewumi2017survey,ball2017comprehensive}).
This review aims to provide a comprehensive outline of state-of-the-art research in DAD techniques as well as several real-world applications these techniques is presented.
\item In recent years several new deep learning based anomaly detection techniques with greatly reduced computational requirements have been developed. The purpose of this paper is to survey these techniques and classify them into an organized schema for better understanding. We introduce two more sub-categories Hybrid models ~(\cite{erfani2016high})and one-class neural networks techniques ~(\cite{chalapathy2018anomaly}) as illustrated in Figure~\ref{fig:surveyTaxonomy} based on the choice of training objective. For each category we discuss both the assumptions and techniques adopted for best performance. Furthermore, within each category, we also present the challenges, advantages, and disadvantages and provide an overview of the computational complexity of DAD methods.
\end{itemize}

\section{Organization}
This chapter is organized by following structure described in Figure~\ref{fig:surveyTaxonomy}.
In Section~\ref{sec:aspectsOfAnomalyDetection}, we identify the various aspects that determine the formulation of the problem and highlight the richness and complexity associated with anomaly detection.
We introduce and define two types of models: contextual and collective or group anomalies. In Section~\ref{sec:applicationsOfDLAD}, we briefly describe the different application domains to which deep learning-based anomaly detection has been applied. In subsequent sections, we provide a categorization of deep learning-based techniques based on the research area to which they belong. 
Based on training objectives employed and availability of labels deep learning-based anomaly detection techniques can be categorized into supervised (Section~\ref{sec:supervisedDAD}), unsupervised (Section ~\ref{sec:unsupervisedDAD}), hybrid (Section~\ref{sec:hybridModels}), and one-class neural network (Section~\ref{sec:oneclassNN}). For each category of techniques we also discuss their computational complexity for training and testing phases. In Section~\ref{sec:typeBasedAD} we discuss the point, contextual, and collective (group) deep learning-based anomaly detection techniques. We present some discussion of the limitations and relative performance of various existing techniques in Section~\ref{sec:relativeSOW}. Section~\ref{sec:chapter1_conclusion} contains concluding remarks.


\section{Different aspects of deep learning-based anomaly detection. }
\label{sec:aspectsOfAnomalyDetection}
This section identifies and discusses the different aspects of deep learning-based anomaly detection.

\subsection{ Nature of Input Data}
The choice of a deep neural network architecture in deep anomaly detection methods primarily depends on the nature of input data. Input data can be broadly classified into sequential (eg, voice, text, music, time series, protein sequences) or non-sequential data (eg, images, other data). Table~\ref{tab:dataTypeModelArchitecture} illustrates the nature of input data and deep model architectures used in anomaly detection. Additionally input data depending on the number of features (or attributes) can be further classified into either low or high-dimensional data. DAD techniques have been to learn complex hierarchical feature relations within high-dimensional raw input data ~(\cite{lecun2015deep}). The number of layers used in DAD techniques is driven by input data dimension, deeper networks are shown to produce better performance on high dimensional data. Later on, in Section ~\ref{sec:deepDADModels} various models considered for outlier detection are reviewed at depth.
\subsection{Based on Availability of labels}
Labels indicate whether a chosen data instance is normal or an outlier. Anomalies are rare entities hence it is challenging to obtain their labels. Furthermore, anomalous behavior may change over time, for instance, the nature of anomaly had changed so significantly and that it remained unnoticed at Maroochy water treatment plant, for a long time which resulted in leakage of 150 million liters of untreated sewerage to local waterways ~(\cite{ramotsoela2018survey}).\\
Deep anomaly detection (DAD) models can be broadly classified into three categories based on the extent of availability of labels. (1) Supervised deep anomaly detection. (2) Semi-supervised deep anomaly detection. (3) Unsupervised deep anomaly detection.

\subsubsection{Supervised deep anomaly detection}
\label{supervised_learning}
Supervised deep anomaly detection involves training a deep supervised binary or multi-class classifier, using labels of both normal and anomalous data instances. For instance supervised DAD models, formulated as multi-class classifier aids in detecting rare brands, prohibited drug name mention and fraudulent health-care transactions ~(\cite{chalapathy2016investigation,chalapathy2016bidirectional}). Despite the improved performance of supervised DAD methods, these methods are not as popular as semi-supervised or unsupervised methods, owing to the lack of availability of labeled training samples. Moreover, the performance of deep supervised classifier used an anomaly detector is sub-optimal due to class imbalance (the total number of positive class instances are far more than the total number of negative class of data). Therefore we do not consider the review of supervised DAD methods in this survey.

\begin{table}
\centering
\scalebox{0.99}{
\begin{tabular}{ |c|c|c|c| }
\hline
Type of Data & Examples & DAD model architecture  \\ [0.5ex]
\hline
\multirow{3}{3em}{Sequential} & Video,Speech &  \\
&Protein Sequence,Time Series  & CNN, RNN, LSTM  \\
&Text (Natural language)  &  \\
\hline
\multirow{2}{3em}{Non-Sequential} & Image,Sensor &  \\
&Other (data)  & CNN, AE and its variants  \\
\hline
\end{tabular}}
\caption{Table illustrating nature of input data and corresponding deep anomaly detection model architectures proposed in literature.
        \\CNN: Convolution Neural Networks, LSTM : Long Short Term Memory Networks \\
         AE: Autoencoders. }
\label{tab:dataTypeModelArchitecture}
\end{table}
\subsubsection{Semi-supervised deep anomaly detection}
\label{semi_supervised_learning}
The labels of normal instances are far more easy to obtain than anomalies, as a result, semi-supervised DAD techniques are more widely adopted, these techniques leverage existing labels of single (normally positive class) to separate outliers. One common way of using deep autoencoders in anomaly detection is to train them in a semi-supervised way on data samples with no anomalies. With sufficient training samples, of normal class autoencoders would produce low reconstruction errors for normal instances, over unusual events ~(\cite{wulsin2010semi,nadeem2016semi,song2017hybrid}). We consider a detailed review of these methods in Section~\ref{sec:semi_supervised_DAD}.

\begin{figure}[h]
\centering
\includegraphics[scale=0.7]{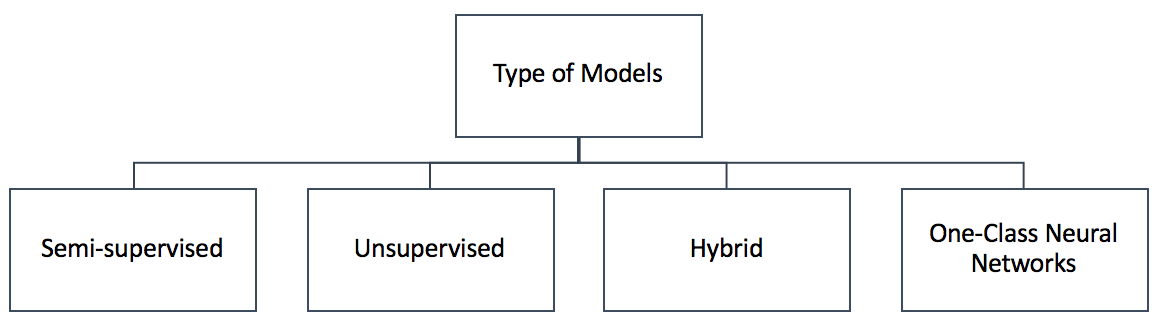}
\captionsetup{justification=centering}
\caption{Taxonomy based on the type of deep learning models for anomaly detection.}
\label{fig:typeOfModels}
\end{figure}

\subsubsection{Unsupervised deep anomaly detection}
\label{sec:USAD}
Unsupervised deep anomaly detection techniques detect outliers solely based on intrinsic properties of the data instances. Unsupervised DAD techniques are used in automatic labeling of unlabelled data samples since labeled data is very hard to obtain ~(\cite{patterson2017deep}). Variants of Unsupervised DAD models~(\cite{tuor2017deep}) are shown to outperform traditional methods such as principal component analysis (PCA) ~(\cite{wold1987principal}), support vector machine (SVM) ~\cite{cortes1995support} and Isolation Forest~(\cite{liu2008isolation}) techniques in applications domains such as health and cyber-security.
Autoencoders are the core of all Unsupervised DAD models. These models assume a high prevalence of normal instances than abnormal data instances failing which would result in high false positive rate. Additionally unsupervised learning algorithms such as restricted Boltzmann machine (RBM)~(\cite{sutskever2009recurrent}), deep Boltzmann machine (DBM), deep belief network (DBN)~(\cite{salakhutdinov2010efficient}), generalized denoising autoencoders~(\cite{vincent2008extracting}) , recurrent neural network (RNN)~(\cite{rodriguez1999recurrent}) Long short term memory networks~(\cite{lample2016neural}) which are used to detect outliers are discussed in detail in Section ~\ref{sec:rnn_lstm_gru}.

\subsection{Based on the training objective}
In this survey we introduce two new categories of deep anomaly detection (DAD) techniques based on training objectives employed 1) Deep hybrid models (DHM). 2) One class neural networks (OC-NN).

\subsubsection{Deep Hybrid Models (DHM)}
\label{sec:DHM}

\begin{figure*}
\centering
\includegraphics[scale=0.7]{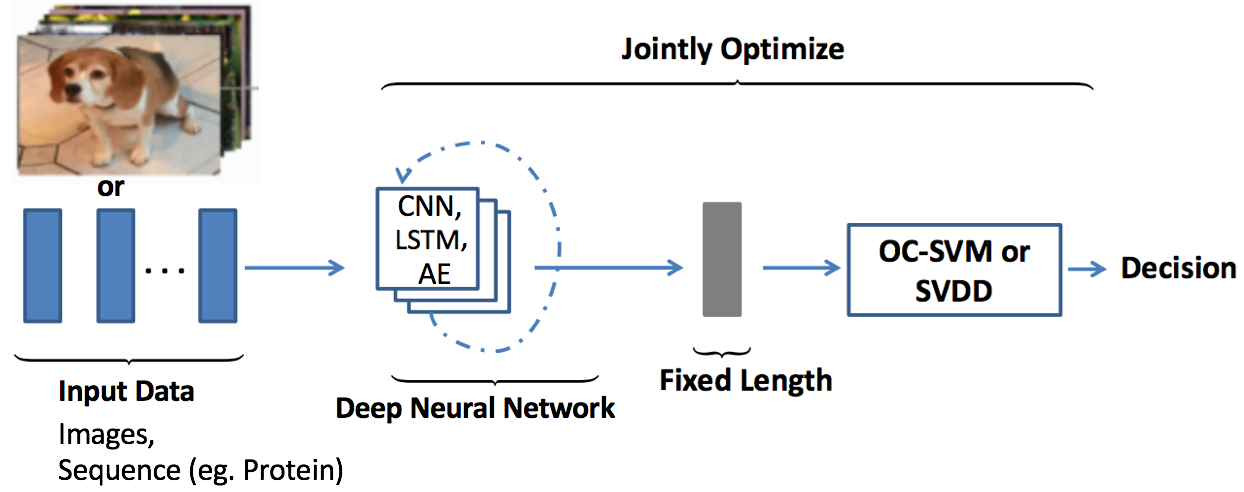}
\captionsetup{justification=centering}
\caption{Deep Hybrid Model Architecture.}
\label{fig:HybridDeepModels}
\end{figure*}

Deep hybrid models for anomaly detection use deep neural networks mainly autoencoders as feature extractors, the features learned within the hidden representations of autoencoders are input to traditional anomaly detection algorithms such as one-class SVM (OC-SVM) to detect outliers~(\cite{andrews2016detecting}). Figure~\ref{fig:HybridDeepModels} illustrates  the deep hybrid model architecture used for anomaly detection. Following the success of transfer learning to obtain rich representative features  from models pre-trained on large data-sets,  hybrid models have also employed these pre-trained transfer learning models as feature extractors with great success ~(\cite{pan2010survey}). A variant of hybrid model was proposed by ~\cite{ergen2017unsupervised} which considers joint training of feature extractor along-with OC-SVM (or SVDD) objective to maximize the detection performance. A notable shortcoming of these hybrid approaches is the lack of trainable objective customized for anomaly detection, hence these models fail to extract rich differential features to detect outliers. In order to overcome this limitation customized objective for anomaly detection such as Deep one-class classification ~(\cite{ruff2018deep}) and One class neural networks ~(\cite{chalapathy2018anomaly}) is introduced.

\subsubsection{One-Class Neural Networks (OC-NN)}
\label{sec:oc-nn}

One class neural network (OC-NN)~\cite{chalapathy2018anomaly} methods are inspired by kernel-based one-class classification which combines the ability of deep networks to extract a progressively rich representation of data with the one-class objective of creating a tight envelope around normal data. The OC-NN approach breaks new ground for the following crucial reason: data representation in the hidden layer is driven by the OC-NN objective and is thus customized for anomaly detection. This is a departure from other approaches which use a hybrid approach of learning deep features using an autoencoder and then feeding the features into a separate anomaly detection method like one-class SVM (OC-SVM).  The details of training and evaluation of one class neural networks is discussed in Section ~\ref{sec:oneclassNN}. Another variant of one class neural network architecture Deep Support Vector Data Description (Deep SVDD)~(\cite{ruff2018deep}) trains deep neural network to extract common factors of variation by closely mapping the normal data instances to the center of sphere, is shown to produce performance improvements on MNIST~(\cite{lecun2010mnist}) and CIFAR-10 ~(\cite{krizhevsky2009learning}) datasets.


\subsection{Type of Anomaly}
\label{sec:typeBasedAD}
Anomalies can be broadly  classified into three types: point anomalies, contextual anomalies and collective anomalies. Deep anomaly detection (DAD) methods have been shown to detect all three types of anomalies with great success.

\begin{figure}[h]
\centering
\includegraphics[scale=0.7]{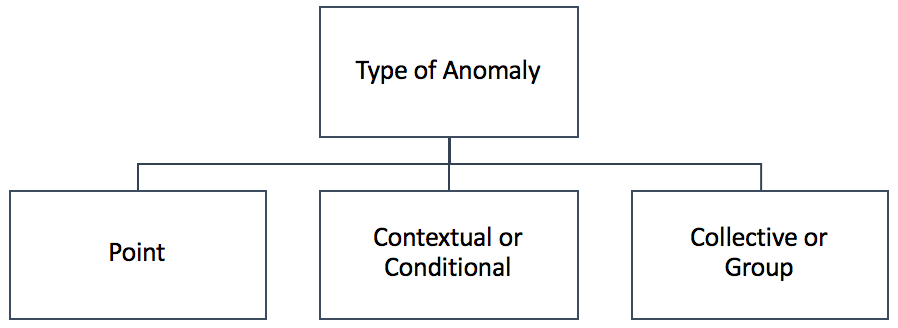}
\captionsetup{justification=centering}
\caption{Deep learning techniques classification based on the type of anomaly.}
\label{fig:typeOfAnomaly}
\end{figure}

\subsubsection{Point Anomalies}
\label{sec:PointAnomalies}
The majority of work in literature focuses on point anomalies. Point anomalies often represent an irregularity or deviation that happens randomly and may have no particular interpretation. For instance, in Figure~\ref{fig:PointAndCollectiveAnomaly} a credit card transaction with high expenditure recorded at
\textit{Monaco} restaurant seems a point anomaly since it significantly deviates from the rest of the transactions. Several real world applications, considering point anomaly detection, are reviewed in Section~\ref{sec:applicationsOfDLAD}.

\subsubsection{Contextual Anomaly Detection}
\label{sec:contextualanomalies}
\begin{figure}[htp]
  \centering
  \subcaptionbox{ Temperature data~\cite{hayes2015contextual}.\label{fig:2dtemp}}{\includegraphics[scale=0.35]{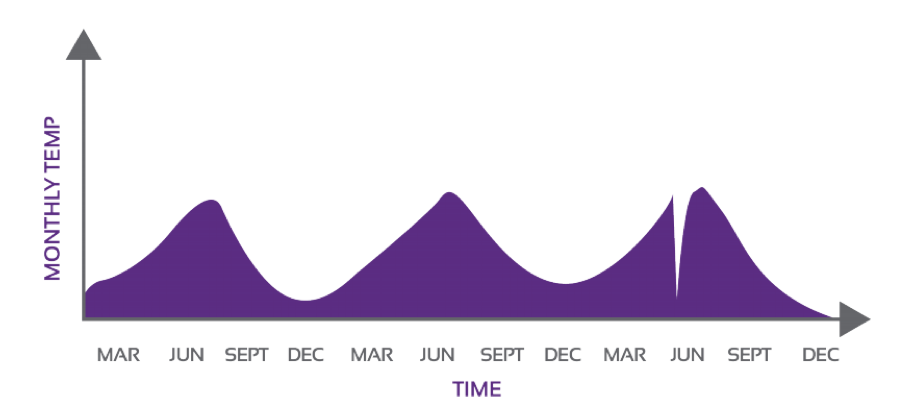}}\hspace{1em}%
  \subcaptionbox{System logs ~\cite{du2017deeplog}.\label{fig:syslog}}{\includegraphics[scale=0.35]{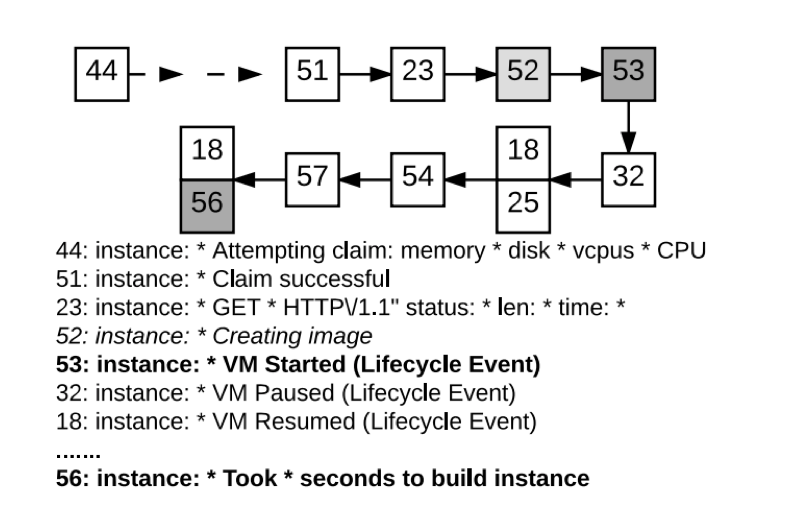}}
  \caption{Illustration of contextual anomaly detection.}
        \label{fig:deeplogContextual}
\end{figure}

A contextual anomaly is also known as the conditional anomaly is a data instance that could be considered as anomalous in some specific context ~(\cite{song2007conditional}). Contextual anomaly is identified by considering both contextual and behavioural features.
 The contextual features, normally used are time and space. While the behavioral features may be a pattern of spending money, the occurrence of system log events or any feature used to describe the normal behavior. Figure ~\ref{fig:2dtemp} illustrates the example of a contextual anomaly considering temperature data indicated by a drastic drop just before June; this value is not indicative of a normal value found during this time. Figure ~\ref{fig:syslog} illustrates using deep Long Short-Term Memory (LSTM) ~(\cite{hochreiter1997long}) based model to identify anomalous system log events ~(\cite{du2017deeplog}) in a given context (e.g event 53 is detected as being out of context).


\subsubsection{Collective or Group Anomaly Detection.}
\label{sec:groupanomaly}
\begin{figure}[h]
\centering
\includegraphics[scale=0.5]{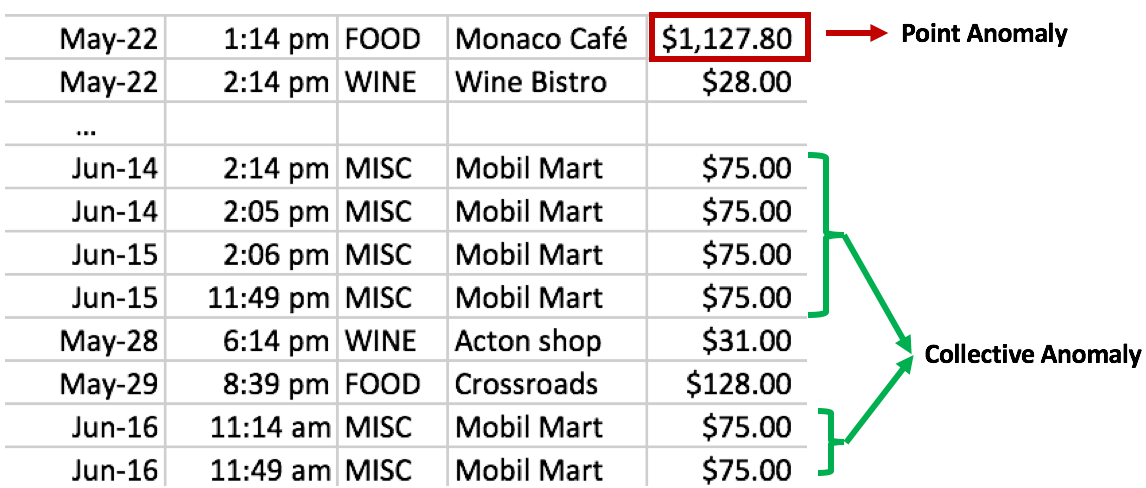}
\captionsetup{justification=centering}
\caption{Credit Card Fraud Detection: Illustrating Point and Collective anomaly.}
\label{fig:PointAndCollectiveAnomaly}
\end{figure}

Anomalous collections of individual data points are known as collective or group anomalies, wherein each of the individual points in isolation appears as normal data instances while observed in a group exhibit unusual characteristics. For example, consider an illustration of a fraudulent credit card transaction, in the log data shown in Figure~\ref{fig:PointAndCollectiveAnomaly}, if a single transaction of "MISC" would have occurred, it might probably not seem as anomalous. The following group of transactions of valued at $\$75$ certainly seems to be a candidate for collective or group anomaly. Group anomaly detection (GAD) with an emphasis on irregular group distributions (e.g., irregular mixtures of image pixels are detected using a variant of autoencoder model ~(\cite{chalapathy2018group,bontemps2016collective,araya2016collective,zhuang2017group}).

\subsection {Output of DAD Techniques}
\label{output_of_dad_methods}
A critical aspect for anomaly detection methods is the way in which the anomalies are detected. Generally, the outputs produced by anomaly detection methods are either anomaly score or binary labels.

\subsubsection{Anomaly Score:}
 Anomaly score describes the level of \textit{outlierness} for each data point. The data instances may be ranked according to anomalous score, and a domain-specific threshold (commonly known as decision score) will be selected by subject matter expert to identify the anomalies. In general, decision scores reveal more information than binary labels. For instance, in Deep SVDD approach the decision score is the measure of the distance of data point from the center of the sphere, the data points which are farther away from the center are considered anomalous ~(\cite{pmlrv80ruff18a}).
 
\subsubsection{Labels:} 
Instead of assigning scores, some techniques may assign a category label as normal or anomalous to each data instance. Unsupervised anomaly detection techniques using autoencoders measure the magnitude of the residual vector (i,e reconstruction error) for obtaining anomaly scores, later on, the reconstruction errors are either ranked or thresholded by domain experts to label data instances.


\section{Applications of Deep Anomaly Detection}
\label{sec:applicationsOfDLAD}

In this section, we discuss several applications of deep anomaly detection. For each application domain, we discuss the following four aspects:\\
\textemdash the notion of an anomaly;\\
\textemdash nature of the data;\\
\textemdash challenges associated with detecting anomalies;\\
\textemdash existing deep anomaly detection techniques.\\

\subsection{Intrusion Detection}
\label{sec:intrusion_detection}
\begin{figure*}
\centering
\includegraphics[scale=0.4]{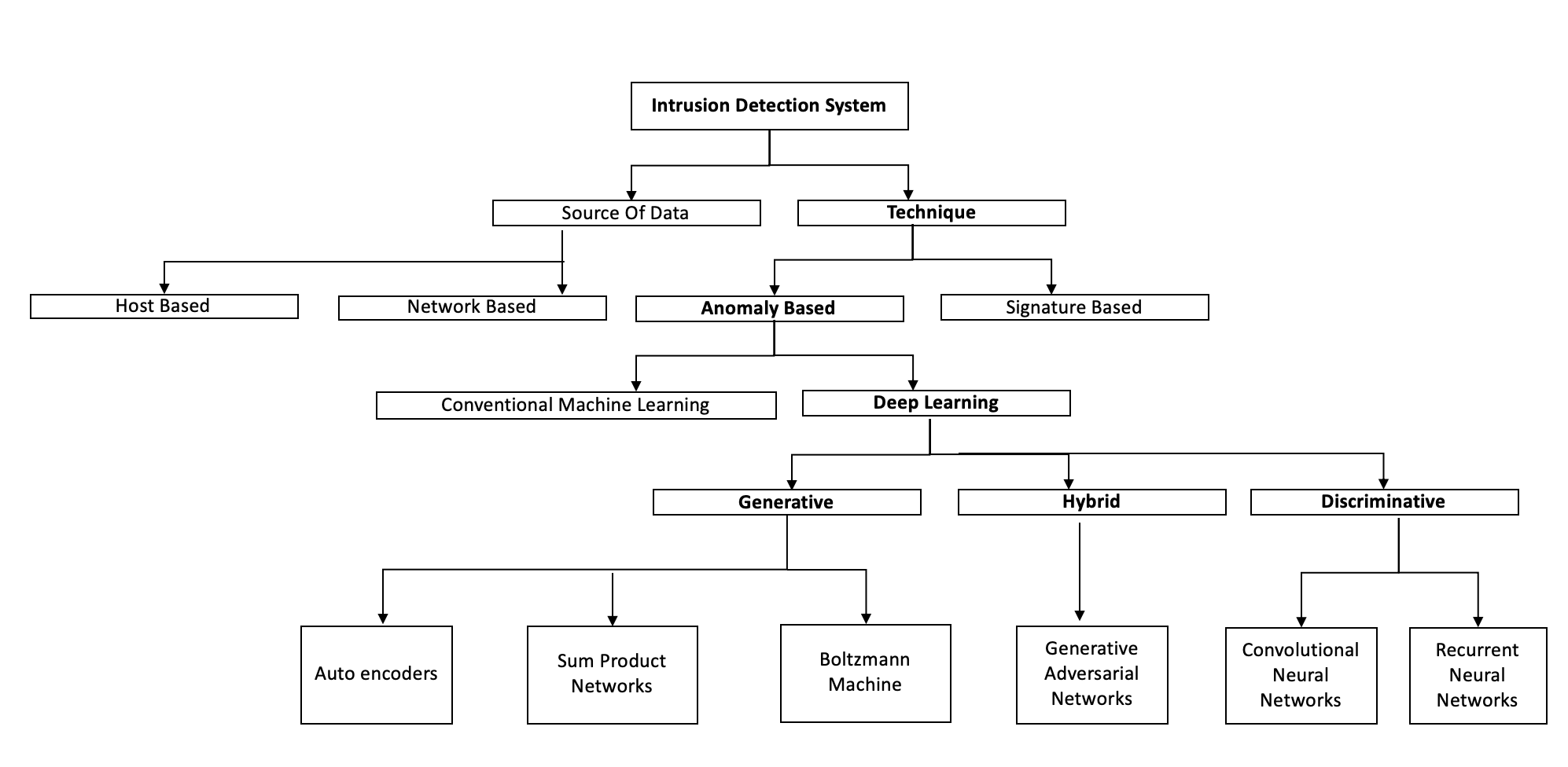}
\captionsetup{justification=centering}
\caption{Classification of deep learning methods for intrusion detection.}
\label{fig:deepADforIDS}
\end{figure*}

The intrusion detection system (IDS) refers to identifying malicious activity in a computer-related system (\cite{phoha2002internet}). IDS may be deployed at single computers known as Host Intrusion Detection (HIDS) to large networks Network Intrusion Detection (NIDS). The classification of deep anomaly detection techniques for intrusion detection is in Figure ~\ref{fig:deepADforIDS}. IDS depending on detection method are classified into signature-based or anomaly based. Using signature-based IDS is not efficient to detect new attacks, for which no specific signature pattern is available, hence anomaly based detection methods are more popular. In this survey, we focus on deep anomaly detection (DAD) methods and architectures employed in intrusion detection.

\vspace{-0.3cm}
\subsubsection{Host-Based Intrusion Detection Systems (HIDS):}
 Such systems are installed software programs which monitors a single host or computer for malicious activity or policy violations by listening to system calls or events occurring within that host (\cite{vigna2005host}). The system call logs could be generated by programs or by user interaction resulting in logs as shown in Figure ~\ref{fig:syslog}. Malicious interactions lead to the execution of these system calls in different sequences. HIDS may also monitor the state of a system, its stored information, in Random Access Memory (RAM), in the file system, log files or elsewhere for a valid sequence. Deep anomaly detection (DAD) techniques applied for HIDS are required to handle the variable length and sequential nature of data. The DAD techniques have to either model the sequence data or compute the similarity between sequences. Some of the success-full DAD techniques for HIDS is illustrated in Table~\ref{tab:HIDS}.

\subsubsection{Network Intrusion Detection Systems (NIDS):}
NIDS systems deal with monitoring the entire network for suspicious traffic by examining each and every network packet. Owing to real-time streaming behavior, the nature of data is synonymous to big data with high volume, velocity, variety. The network data also has a temporal aspect associated with it. Some of the success-full DAD techniques for NIDS is illustrated in Table~\ref{tab:NIDS} . This survey also lists the data-sets used for evaluating the DAD intrusion detection methods in Table~\ref{tab:IDSDataset}. A challenge faced by DAD techniques in intrusion detection is that the nature of anomalies keeps changing over time as the intruders adapt their network attacks to evade the existing intrusion detection solutions.

\begin{table}
\begin{center}
\caption{Examples of DAD Techniques employed in HIDS
          \\CNN: Convolution Neural Networks, LSTM : Long Short Term Memory Networks
          \\GRU: Gated Recurrent Unit, DNN : Deep Neural Networks
          \\SPN: Sum Product Networks}
 \captionsetup{justification=centering}
  \label{tab:HIDS}
   \scalebox{0.82}{
    \begin{tabular}{ | l | p{4cm} | l | p{5cm} |}
    \hline
    Techniques & Model Architecture & Section & References \\ \hline
    Discriminative &  LSTM , CNN-LSTM-GRU, DNN & Section ~\ref{sec:rnn_lstm_gru},~\ref{sec:cnn},~\ref{sec:dnn} &  \cite{kim2016lstm},\cite{chawla2018host},\cite{chen2018henet},\cite{sohi2018recurrent},\cite{vinayakumar2017applying} \\\hline
    Hybrid &  GAN & Section ~\ref{sec:hybridModels} & \cite{aghakhani2018detecting}, \cite{li2018anomaly} \\\hline
    Generative &  AE, SPN,  & Section ~\ref{sec:ae},~\ref{sec:spn} & \cite{gao2014intrusion},\cite{peharz2018probabilistic},\cite{umer2018two} \\
    \hline
    \end{tabular}}
\end{center}
\end{table}

\begin{table}
\begin{center}
  \caption{Examples of DAD Techniques employed in NIDS.
          \\CNN: Convolution Neural Networks, LSTM : Long Short Term Memory Networks
          \\RNN: Recurrent Neural Networks, RBM : Restricted Boltzmann Machines
          \\DCA: Dilated Convolution Autoencoders, DBN : Deep Belief Network
          \\AE: Autoencoders, SAE: Stacked Autoencoders
          \\GAN: Generative Adversarial Networks, CVAE : Convolutional Variational Autoencoder. }
  \captionsetup{justification=centering}

  \label{tab:NIDS}
  \scalebox{0.80}{
    \begin{tabular}{ | l | p{4cm} | p{5cm}  | p{7cm} |}
    \hline
    Techniques & Model Architecture & Section & References \\ \hline
   Generative  & DCA, SAE, RBM, DBN, CVAE & Section ~\ref{sec:cnn},~\ref{sec:ae},~\ref{sec:dnn},~\ref{sec:gan_adversarial} & \cite{yu2017network},\cite{thing2017ieee}, \cite{zolotukhin2016increasing},~\cite{cordero2016analyzing},\cite{alrawashdeh2016toward},\cite{tang2016deep},\cite{lopez2017conditional},\cite{al2018deep},\cite{mirsky2018kitsune},\cite{aygun2017network} \\ \hline
  Hybrid  & GAN   & Section ~\ref{sec:hybridModels} & \cite{lin2018idsgan},\cite{yin2018enhancing}, \cite{ring2018flow}, \cite{latah2018deep},\cite{intrator2018mdgan},\cite{matsubara2018anomaly},~\cite{nicolau2016hybrid} ,\cite{rigaki2017adversarial}. \\ \hline
  Discriminative &  RNN , LSTM ,CNN & Section ~\ref{sec:rnn_lstm_gru},~\ref{sec:cnn} & \cite{yu2017network}, \cite{malaiya2018empirical} \cite{kwon2018empirical},\cite{gao2014intrusion},\cite{staudemeyer2015applying},\cite{naseer2018enhanced}\\
  \hline
  \end{tabular}}
\end{center}
\end{table}

\begin{table*}
\begin{center}
  \caption{Datasets Used in Intrusion Detection }
   \label{tab:IDSDataset}
    \scalebox{0.85}{
    \begin{tabular}{ | p{3cm} | l | p{5cm} | p{3cm} | p{5cm} |}
    \hline
    DataSet &IDS & Description & Type & References \\ \hline
    CTU-UNB & NIDS & CTU-UNB~\cite{ucsdAnomalyDetect2017} dataset consists of various botnet traffics from CTU-13 dataset [20] and normal traffics from the UNB ISCX IDS 2012 dataset ~\cite{shiravi2012toward}  & Hexadecimal & \cite{yu2017network} \\ \hline
    Contagio-CTU-UNB & NIDS  & Contagio-CTU-UNB dataset consists of six types of network traffic data. ~\cite{adam2008robust}  & Text & ~\cite{yu2017network}. \\ \hline
    NSL-KDD~\footnote{http://nsl.cs.unb.ca/NSL-KDD/}& NIDS &The NSL-KDD data set is a refined version of its predecessor KDD-99 data set.  ~\cite{ucsdAnomalyDetect2017} & Text &  ~\cite{yin2017deep},~\cite{javaid2016deep},  ~\cite{tang2016deep},~\cite{yousefi2017autoencoder},~\cite{mohammadi2017new}, ~\cite{lopez2017conditional}\\\hline
    DARPA KDD- CUP 99 & NIDS & DARPA KDD~\cite{stolfo2000cost} The competition task was to build a network intrusion detector, a predictive model capable of distinguishing between ``bad'' connections, called intrusions or attacks, and ``good'' normal connections.  & Text   & ~\cite{alrawashdeh2016toward} ,~\cite{van2017anomaly},~\cite{mohammadi2017new}\\\hline
    MAWI& NIDS  & The  MAWI~\cite{fontugne2010mawilab}  dataset  consists  of  network  traffic  capturedfrom  backbone  links  between  Japan  and  USA.  Every  daysince  2007  & Text   & ~\cite{cordero2016analyzing} \\\hline
    Realistic Global Cyber Environment (RGCE) & NIDS & RGCE~\cite{jamkRGCE}  contains
    realistic Internet Service Providers (ISPs) and numerous different web services as
    in the real Internet.  &  Text   & ~\cite{zolotukhin2016increasing} \\\hline
    ADFA-LD& HIDS & The ADFA Linux Dataset (ADFA-LD). This dataset provides a contemporary Linux dataset for evaluation by traditional HIDS~\cite{creech2014semantic} & Text   & ~\cite{kim2016lstm},~\cite{chawla2018host} \\\hline
    UNM-LPR& HIDS & Consists of system calls to evalute HIDS system~\cite{ImmuneDatasets} & Text   & ~\cite{kim2016lstm} \\\hline
    Infected PDF samples& HIDS & Consists of set of Infected PDF samples, which are used to monitor the malicious traffic  & Text   &~\cite{chen2018henet}\\\hline
    \end{tabular}}
\end{center}
\end{table*}

\label{sec:intrusionDetect}

\subsection{Fraud Detection}

\begin{figure}[h]
\centering
\includegraphics[scale=0.5]{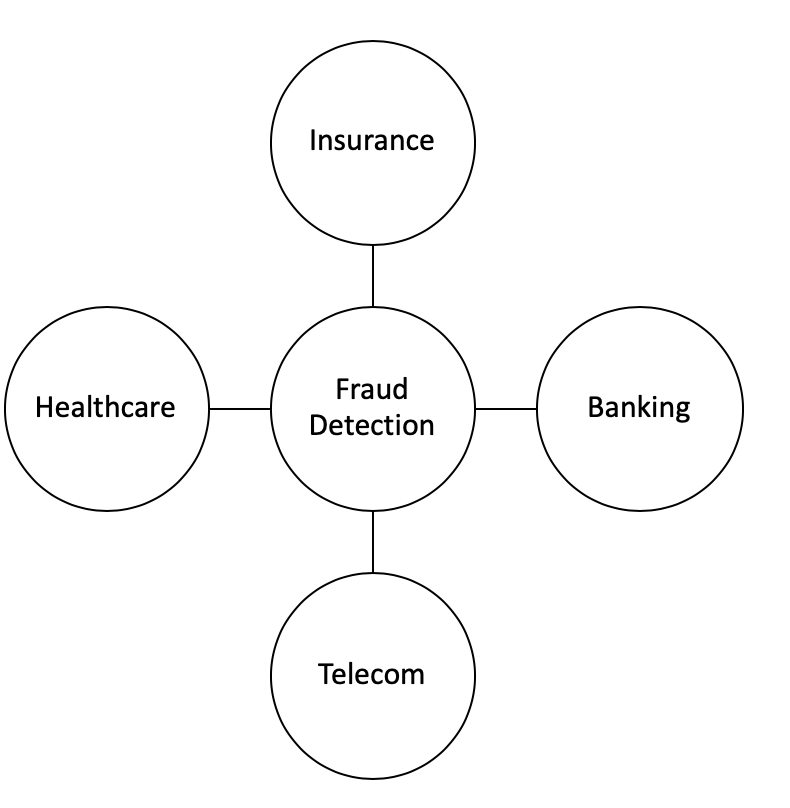}
\captionsetup{justification=centering}
\caption{Fraud detection across various application domains.}
\label{fig:AerasOfFraud}
\end{figure}

Fraud is a deliberate act of deception to access valuable resources ~(\cite{abdallah2016fraud}). The PricewaterhouseCoopers (PwC) global economic crime survey of 2018 ~(\cite{Lavion2018,zhao2013fraud}) found that half of the 7,200 companies they surveyed had experienced fraud of some nature. Fraud detection refers to the detection of unlawful activities across various industries, illustrated in ~\ref{fig:AerasOfFraud}.

Fraud in telecommunications, insurance \textit{( health, automobile, etc)} claims, banking \textit{( tax return claims, credit card transactions etc)} represent significant problems in both governments and private businesses. Detecting and preventing fraud is not a simple task since fraud is an adaptive crime. Many traditional machine learning algorithms have been applied successfully in fraud detection ~(\cite{sorournejad2016survey}). The challenge associated with detecting fraud is that it requires real-time detection and prevention. This section focuses on deep anomaly detection (DAD) techniques for fraud detection.


\subsubsection{Banking fraud}
Credit card has become a popular payment method in online shopping for goods and services. Credit card fraud involves theft of a payment card details, and use it as a fraudulent source of funds in a transaction. Many techniques for credit card fraud detection have been presented in the last few years ~(\cite{zhou2018state,suganya2015survey}). We will briefly review some of DAD techniques as shown in Table~\ref{tab:creditfraudDetect}. The challenge in credit card fraud detection is that frauds have no consistent patterns. The typical approach in credit card fraud detection is to maintain a usage profile for each user and monitor the user profiles to detect any deviations. Since there are billions of credit card users this technique of user profile approach is not very scalable. Owing to the inherent scalable nature of DAD techniques techniques are gaining broad spread adoption in credit card fraud detection.

\begin{table*}
\begin{center}
  \caption{Examples of DAD techniques used in  credit card fraud detection.
          \\AE: Autoencoders, LSTM : Long Short Term Memory Networks
          \\RBM: Restricted Botlzmann Machines, DNN : Deep Neural Networks
          \\GRU: Gated Recurrent Unit, RNN: Recurrent Neural Networks
          \\CNN: Convolutional Neural Networks,VAE: Variational Autoencoders
          \\GAN: Generative Adversarial Networks}
  \captionsetup{justification=centering}
  \label{tab:creditfraudDetect}
  \scalebox{0.83}{
    \begin{tabular}{ | l | p{3cm} | p{10cm} |}
    \hline
    Technique Used & Section & References \\ \hline
     AE  & Section ~\ref{sec:ae}  & ~\cite{schreyer2017detection},~\cite{wedge2017solving} ,~\cite{paula2016deep},~\cite{renstrom2018fraud},~\cite{kazemi2017using},~\cite{zheng2018one},~\cite{pumsirirat2018credit} \\\hline
     RBM  & Section ~\ref{sec:dnn}  & ~\cite{pumsirirat2018credit} \\\hline
     DBN & Section ~\ref{sec:dnn} & ~\cite{seeja2014fraudminer} \\\hline
     VAE & Section ~\ref{sec:gan_adversarial} & ~\cite{sweers2018autoencoding} \\\hline
     GAN & Section ~\ref{sec:gan_adversarial} & ~\cite{fiore2017using},~\cite{choi2018generative} \\\hline
     DNN  & Section ~\ref{sec:dnn} & ~\cite{dorronsoro1997neural}, ~\cite{gomez2018end} \\\hline
     LSTM,RNN,GRU  & Section ~\ref{sec:rnn_lstm_gru} & ~\cite{wiese2009credit}, ~\cite{jurgovsky2018sequence},~\cite{heryadi2017learning},~\cite{ando2016detecting},~\cite{wang2017session},~\cite{alowais2012credit},~\cite{amarasinghe2018critical},~\cite{abroyan2017neural},~\cite{lp2018transaction}\\\hline
     CNN  & Section ~\ref{sec:cnn} & ~\cite{shen2007application},~\cite{chouiekh2018convnets},~\cite{abroyan2017convolutional},~\cite{fu2016credit},~\cite{lu2017deep},~\cite{wang2018credit},~\cite{abroyan2017neural} ,~\cite{zhang2018model}\\\hline
    \end{tabular}}
\end{center}
\end{table*}

\subsubsection{Mobile cellular network fraud}
\label{sec:mobilefraud}

In recent times, mobile cellular networks have witnessed rapid deployment and evolution supporting billions of users and a vastly diverse array of mobile devices. Due to this broad adoption and low mobile cellular service rates, mobile cellular networks is now faced with frauds such as voice scams targeted to steal customer private information, and messaging related scams to extort money from customers. Detecting such fraud is of paramount interest and not an easy task due to volume and velocity of the mobile cellular network. Traditional machine learning methods with static feature engineering techniques fail to adapt to the nature of evolving fraud. Table ~\ref{tab:mobilefraudDetect} lists DAD techniques for mobile cellular network fraud detection.

\begin{table*}
\begin{center}
  \caption{Examples of DAD techniques used in  mobile cellular network fraud detection.
          \\CNN:  convolution neural networks,DBN: Deep Belief Networks
          \\SAE: Stacked Autoencoders, DNN : Deep neural networks
          \\GAN: Generative Adversarial Networks }
  \captionsetup{justification=centering}
  \label{tab:mobilefraudDetect}
  \scalebox{0.9}{
    \begin{tabular}{ | l | p{4cm} | p{5cm} | p{5cm} |}
    \hline
    Technique Used & Section & References \\ \hline
     CNN    & Section ~\ref{sec:cnn}  & ~\cite{chouiekh2018convnets} \\\hline
     SAE, DBN    & Section \ref{sec:ae},~\ref{sec:dnn}   &  ~\cite{alsheikh2016mobile},~\cite{badhe2017click} \\\hline
     DNN & Section ~\ref{sec:dnn}  &  ~\cite{akhter2012detecting},~\cite{jain2017perspective} \\\hline
     GAN & Section ~\ref{sec:gan_adversarial} &  ~\cite{zheng2018generative} \\\hline
    \end{tabular}}
\end{center}
\end{table*}
\subsubsection{Insurance fraud}

Several traditional machine learning methods have been applied successfully to detect fraud in insurance claims ~(\cite{joudaki2015using,roy2017detecting}). The traditional approach for fraud detection is based on features which are fraud indicators. The challenge with these traditional approaches is that the need for manual expertise to extract robust features. Another challenge is insurance fraud detection is the that the incidence of frauds is far less than the total number of claims, and also each fraud is unique in its way. In order to overcome these limitations several DAD techniques are proposed which are illustrated in Table ~\ref{tab:insurancefraudDetect}.

\begin{table*}
\begin{center}
  \caption{Examples of DAD techniques used in insurance fraud detection.
          \\DBN: Deep Belief Networks, DNN : Deep Neural Networks
          \\CNN: Convolutional Neural Networks,VAE: Variational Autoencoders
          \\GAN: Generative Adversarial Networks}
  \label{tab:insurancefraudDetect}
  \captionsetup{justification=centering}
  \scalebox{0.85}{
    \begin{tabular}{ | l | p{4cm} | p{4cm} | p{4cm} |}
    \hline
     DBN & Section ~\ref{sec:dnn} & ~\cite{viaene2005auto} \\\hline
     VAE & Section ~\ref{sec:gan_adversarial} & ~\cite{fajardo2018vos} \\\hline
     GAN & Section ~\ref{sec:gan_adversarial} & ~\cite{fiore2017using},~\cite{choi2018generative} \\\hline
     DNN & Section ~\ref{sec:dnn} &~\cite{keung2009neural}\\\hline
     CNN  & Section ~\ref{sec:cnn} & ~\cite{shen2007application},~\cite{zhang2018model}\\\hline
    \end{tabular}}
\end{center}
\end{table*}
\subsubsection{Healthcare fraud}

Healthcare is an integral component in people's lives, waste, abuse, and fraud drive up costs in healthcare by tens of billions of dollars each year. Healthcare insurance claims fraud is a significant contributor to increased healthcare costs, but its impact can be mitigated through fraud detection. Several machine learning models have been used effectively in health care insurance fraud ~(\cite{bauder2017medicare}). Table~\ref{tab:healthcarefraudDetect} presents an overview of DAD methods for health-care fraud identification.

\begin{table*}
\begin{center}
  \caption{Examples of DAD techniques used in  healthcare fraud detection.
          \\RBM: Restricted Botlzmann Machines, GAN: Generative Adversarial Networks}
  \captionsetup{justification=centering}
  \label{tab:healthcarefraudDetect}
  \scalebox{0.9}{
    \begin{tabular}{ | l | p{4cm} | p{5cm} | p{5cm} |}
    \hline
    Technique Used & Section & References \\ \hline
     RBM & Section ~\ref{sec:dnn} & ~\cite{lasaga2018deep} \\\hline
     GAN & Section ~\ref{sec:gan_adversarial} & ~\cite{ghasedi2018semi},~\cite{finlayson2018adversarial}\\\hline
     CNN  & Section ~\ref{sec:cnn} & ~\cite{esteva2017dermatologist}\\\hline
    \end{tabular}}
\end{center}
\end{table*}

\label{sec:fraudDetection}

\subsection{Malware Detection}
\label{sec:malwaredetection}
Malware, short for Malicious Software. In order to protect legitimate users from malware, machine learning based efficient malware detection methods are proposed ~(\cite{ye2017survey}). In classical machine learning methods, the process of malware detection is usually divided into two stages: feature extraction and classification/clustering. The performance of traditional malware detection approaches critically depend on the extracted features and the methods for classification/clustering. The challenge associated in malware detection problems is the sheer scale of data, for instance considering data as bytes a specific sequence classification problem could be of the order of two million time steps. Furthermore, the malware is very adaptive in nature, wherein the attackers would use advanced techniques to hide the malicious behavior. Some DAD techniques which address these challenges effectively and detect malware are shown in Table~\ref{tab:malwareDetect}.

\begin{table*}
  \begin{center}
   \caption{Examples of DAD techniques used for malware detection.
             \\AE: Autoencoders, LSTM : Long Short Term Memory Networks
             \\RBM: Restricted Botlzmann Machines, DNN : Deep Neural Networks
             \\GRU: Gated Recurrent Unit, RNN: Recurrent Neural Networks
             \\CNN: Convolutional Neural Networks,VAE: Variational Autoencoders
             \\GAN: Generative Adversarial Networks,CNN-BiLSTM: CNN- Bidirectional LSTM}
    \captionsetup{justification=centering}
    \label{tab:malwareDetect}
    \scalebox{0.85}{
    \begin{tabular}{|p{3cm}|p{2cm}|p{10cm}|}
      \hline
      \textbf{Technique Used} & \textbf{Section} & \textbf{References}\\
      \hline
      AE &  Section ~\ref{sec:ae}  & ~\cite{yousefi2017autoencoder},~\cite{hardy2016dl4md},~\cite{yousefi2017autoencoder},~\cite{de2018malware},~\cite{sewak2018investigation},~\cite{kebede2017classification},~\cite{de2018malware},~\cite{david2015deepsign}\\\hline
      word2vec & Section ~\ref{sec:word2vec} &  ~\cite{cakir2018malware},~\cite{silva2018improving}\\\hline
      CNN & Section ~\ref{sec:cnn} &  ~\cite{kolosnjaji2018adversarial},~\cite{suciu2018exploring},~\cite{srisakaokul2018muldef},~\cite{srisakaokul2018muldef},~\cite{king2018artificial},~\cite{huang2017r2},~\cite{guo2017malware},~\cite{abdelsalam2018malware},\newline ~\cite{raff2017malware},~\cite{karbab2018maldozer},~\cite{martinelli2017evaluating},~\cite{mclaughlin2017deep},~\cite{gibert2018using},~\cite{kolosnjaji2017empowering}\\\hline
      DNN & Section ~\ref{sec:dnn} &  ~\cite{rosenberg2018end},~\cite{wang2017adversary}\\\hline
      DBN & Section ~\ref{sec:dnn} &  ~\cite{david2015deepsign},~\cite{yang2016application},~\cite{ding2016application},~\cite{yuxin2017malware},~\cite{selvaganapathy2018deep},~\cite{yuxin2017malware},~\cite{hou2017deep}\\\hline
      LSTM & Section ~\ref{sec:rnn_lstm_gru} &  ~\cite{tobiyama2016malware}, ~\cite{hu2017black},~\cite{tobiyama2018method} ,~\cite{passalislong} \\\hline
      CNN-BiLSTM& Section ~\ref{sec:cnn},~\ref{sec:rnn_lstm_gru} &  ~\cite{le2018deep},~\cite{wang2017adversary} \\\hline
      GAN& Section ~\ref{sec:gan_adversarial} &  ~\cite{kim2018zero} \\\hline
      Hybrid model(AE-CNN),(AE-DBN) & Section ~\ref{sec:hybridModels} &  ~\cite{wang2018effective},~\cite{li2015hybrid} \\\hline
      RNN & Section ~\ref{sec:rnn_lstm_gru} &  ~\cite{haddadpajouh2018deep} \\\hline
    \end{tabular}}
  \end{center}
\end{table*}


\label{sec:malwareDetection}

\subsection{Medical Anomaly Detection}
\label{sec:medical_anomaly_detection}
Several studies have been conducted to understand the theoretical and practical applications of deep learning in medical and bio-informatics ~(\cite{min2017deep,cao2018deep,zhao2016deep,khan2018review}). Finding rare events (anomalies) in areas such as medical image analysis, clinical electroencephalography (EEG) records, enable to diagnose and provide preventive treatments for a variety of medical conditions. Deep learning based architectures are employed with great success to detect medical anomalies as illustrated in Table ~\ref{tab:medicalanomalyDetect}. The vast amount of imbalanced data in medical domain presents significant challenges to detect outliers. Additionally deep learning techniques for long have been considered as black-box techniques. Even though deep learning models produce outstanding performance, these models lack interpret-ability. In recent times models with good interpret-ability are proposed and shown to produce state-of-the-art performance ~(\cite{gugulothusparse,amarasinghe2018toward,choi2018doctor}).

\begin{table*}
\begin{center}
  \caption{Examples of DAD techniques Used for medical anomaly detection.
          \\AE: Autoencoders, LSTM : Long Short Term Memory Networks
          \\GRU: Gated Recurrent Unit, RNN: Recurrent Neural Networks
          \\CNN: Convolutional Neural Networks,VAE: Variational Autoencoders
          \\GAN: Generative Adversarial Networks, KNN: K-nearest neighbours
          \\RBM: Restricted Boltzmann Machines.}
  \captionsetup{justification=centering}
  \label{tab:medicalanomalyDetect}
   \scalebox{0.9}{
    \begin{tabular}{ | l | p{3cm} | p{9cm} |}
    \hline
    Technique Used & Section & References \\ \hline
     AE  & Section~\ref{sec:ae} & ~\cite{wang2016research,cowton2018combined},~\cite{sato2018primitive}\\\hline
     DBN & Section~\ref{sec:dnn} & ~\cite{turner2014deep},~\cite{sharma2016abnormality},~\cite{wulsin2010semi},~\cite{ma2018unsupervised},~\cite{zhang2016automatic},~\cite{wulsin2011modeling} ,~\cite{wu2015adaptive}\\\hline
     RBM & Section~\ref{sec:dnn}  & ~\cite{liao2016enhanced}\\\hline
     VAE & Section~\ref{sec:gan_adversarial} & ~\cite{xu2018unsupervised},~\cite{lu2018anomaly} \\\hline
     GAN & Section~\ref{sec:gan_adversarial}&~\cite{ghasedi2018semi},~\cite{chen2018unsupervised} \\\hline
     LSTM ,RNN,GRU & Section~\ref{sec:rnn_lstm_gru} & ~\cite{yang2018toward},~\cite{jagannatha2016bidirectional},~\cite{cowton2018combined},~\cite{o2016recurrent},~\cite{latif2018phonocardiographic},~\cite{zhang2018time},~\cite{chauhan2015anomaly},~\cite{gugulothusparse,amarasinghe2018toward}\\\hline
     CNN  & Section~\ref{sec:cnn} & ~\cite{schmidt2018artificial},~\cite{esteva2017dermatologist},~\cite{wang2016research},~\cite{iakovidis2018detecting}\\\hline
     Hybrid( AE+ KNN) & Section~\ref{sec:cnn} & ~\cite{song2017hybrid} \\\hline
    \end{tabular}}
\end{center}
\end{table*}

\label{sec:medicalAnomalyDetect}

\subsection{Deep learning for Anomaly detection in Social Networks}

In recent times, online social networks have become part and parcel of daily life. Anomalies in a social network are irregular often unlawful behavior pattern of individuals within a social network; such individuals may be identified as spammers, sexual predators, online fraudsters, fake users or rumor-mongers. Detecting these irregular patterns is of prime importance since if not detected, the act of such individuals can have a serious social impact. A survey of traditional anomaly detection techniques and its challenges to detect anomalies in social networks is a well studied topic in literature~(\cite{liu2017social,savage2014anomaly,anand2017anomaly,yu2016survey,cao2018automatic,yu2016survey}). The heterogeneous and dynamic nature of data presents significant challenges to DAD techniques. Despite these challenges, several DAD techniques illustrated in Table ~\ref{tab:socialNetworkAnomalyDetect} are shown outperform state-of-the-art methods.

\begin{table*}
  \begin{center}
   \caption{Examples of DAD techniques used to detect anomalies in social network.
            \\CNN: Convolution Neural Networks, LSTM : Long Short Term Memory Networks
            \\AE: Autoencoders, DAE: Denoising Autoencoders
            \\SVM : Support Vector Machines., DNN : Deep Neural Network  }
    \captionsetup{justification=centering}
    \label{tab:socialNetworkAnomalyDetect}
    \scalebox{0.85}{
    \begin{tabular}{|p{3cm}|p{4cm}|p{5cm}|}
      \hline
      \textbf{Technique Used} & \textbf{Section} & \textbf{References}\\
      \hline
      AE,DAE &  Section ~\ref{sec:ae}  & ~\cite{zhang2017detecting},~\cite{castellini2017fake}\\\hline
      CNN-LSTM & Section ~\ref{sec:cnn}, ~\ref{sec:rnn_lstm_gru} & ~\cite{sun2018detecting},~\cite{shu2017doc},~\cite{yang2018anomaly}\\\hline
      DNN & Section ~\ref{sec:dnn}  & ~\cite{li2017detecting}\\\hline
      Hybrid Models (CNN-LSTM-SVM) & Section ~\ref{sec:hybridModels}  & ~\cite{wei2017new}\\\hline
    \end{tabular}}
  \end{center}
\end{table*}


\label{sec:socialNetworks}

\subsection{Log Anomaly Detection}

Anomaly detection in log file aims to find text, which can indicate the reasons and the nature of the failure of a system. Most commonly, a domain-specific regular-expression is constructed from past experience which finds new faults by pattern matching. The limitation of such approaches is that newer messages of failures are easily are not detected~(\cite{memon2008log}). The unstructured and diversity in both format and semantics of log data pose significant challenges to log anomaly detection. Anomaly detection techniques should adapt to the concurrent set of log data generated and detect outliers in real time. Following the success of deep neural networks in real time text analysis, several DAD techniques illustrated in Table~\ref{tab:logAnomalyDetect}  model the log data as a natural language sequence are shown very effective in detecting outliers.

\begin{table*}
\begin{center}
\caption{Examples of Deep learning anomaly detection techniques used in system logs.
        \\CNN: Convolution Neural Networks, LSTM : Long Short Term Memory Networks
        \\GRU: Gated Recurrent Unit, DNN : Deep Neural Networks
        \\AE: Autoencoders, DAE: Denoising Autoencoders}
    \captionsetup{justification=centering}
  \label{tab:logAnomalyDetect}
  \scalebox{0.85}{
    \begin{tabular}{ | p{2cm} | p{2cm} | p{9cm} |}
    \hline
     \textbf{Techniques}  & \textbf{Section} & \textbf{References} \\ \hline
     LSTM & Section ~\ref{sec:rnn_lstm_gru} & ~\cite{hochreiter1997long},~\cite{brown2018recurrent},~\cite{tuor2017deep},~\cite{das2018desh},~\cite{malhotra2015long} \\\hline
     AE & Section ~\ref{sec:ae} & ~\cite{du2017deeplog},~\cite{andrews2016detecting} ,~\cite{sakurada2014anomaly},~\cite{nolle2018analyzing},~\cite{nolle2016unsupervised}\\\hline
     LSTM-AE & Section ~\ref{sec:rnn_lstm_gru}, ~\ref{sec:ae} & ~\cite{grover2018anomaly},~\cite{wolpher2018anomaly} \\\hline
     RNN & Section ~\ref{sec:rnn_lstm_gru} & ~\cite{brown2018recurrent},~\cite{zhang2018role},~\cite{nanduri2016anomaly},~\cite{fengming2017anomaly}\\\hline
     DAE & Section ~\ref{sec:ae} & ~\cite{marchi2015non},~\cite{nolle2016unsupervised}\\\hline
     CNN & Section ~\ref{sec:cnn} & ~\cite{lu2018detecting},~\cite{yuan2018insider},~\cite{racki2018compact},~\cite{zhou2016spatial},~\cite{gorokhov2017convolutional},~\cite{liao2017deep},~\cite{cheng2017deep},~\cite{zhang2018alphamex}\\\hline
    \end{tabular}}
\end{center}
\end{table*}

\label{sec:logAnomaly}

\subsection{Internet of things (IoT) Big Data Anomaly Detection}

IoT is identified as a network of devices that are interconnected with soft-wares, servers, sensors and etc. In the field of the Internet of things (IoT), data generated by weather stations, Radio-frequency identification (RFID) tags, IT infrastructure components, and some other sensors are mostly time-series sequential data. Anomaly detection in these IoT networks identifies fraudulent, faulty behavior of these massive scales of interconnected devices.  The challenges associated with outlier detection is that heterogeneous devices are interconnected which renders the system more complex. A thorough overview of using deep learning (DL), to facilitate analytics and learning in the IoT domain is presented by ~(\cite{mohammadi2018deep}). Table ~\ref{tab:iotBigDataAnomalyDetect} illustrates the DAD techniques employed IoT devices.

\begin{table*}
\begin{center}
\caption{Examples of DAD techniques used in Internet of things (IoT) Big Data Anomaly Detection.
        \\ AE: Autoencoders, LSTM : Long Short Term Memory Networks
        \\ DBN : Deep Belief Networks.}
 \captionsetup{justification=centering}
  \label{tab:iotBigDataAnomalyDetect}
  \scalebox{0.95}{
    \begin{tabular}{ | l | p{4cm} | p{5cm} |}
    \hline
     \textbf{Techniques}  & \textbf{Section} & \textbf{References} \\ \hline
     AE & Section ~\ref{sec:ae} & ~\cite{luo2018distributed},~\cite{mohammadi2018neural} \\\hline
     DBN & Section ~\ref{sec:dnn} & ~\cite{kakanakova2017outlier} \\ \hline
     LSTM & Section ~\ref{sec:rnn_lstm_gru} & ~\cite{zhang2018lstm},~\cite{mudassar2018unsupervised}\\ \hline
    \end{tabular}}
\end{center}
\end{table*}

\label{sec:iotBigDataAnomaly}

\subsection{Industrial Anomalies Detection}

Industrial systems consisting of wind turbines, power plants, high-temperature energy systems, storage devices and with rotating mechanical parts are exposed to enormous stress on a day-to-day basis. Damage to these type of systems not only causes economic loss but also a loss of reputation, therefore detecting and repairing them early is of utmost importance. Several machine learning techniques have been used to detect such damage in industrial systems ~(\cite{ramotsoela2018survey,marti2015anomaly}). Several papers published utilizing deep learning models for detecting early industrial damage show great promise ~(\cite{atha2018evaluation,de2018automatic,wang2018residential}). Damages caused to equipment are rare events, thus detecting such events can be formulated as an outlier detection problem. The challenges associated with outlier detection in this domain is both volumes as well as the dynamic nature of data since failure is caused due to a variety of factors. Some of the DAD techniques employed across various industries are illustrated in Table ~\ref{tab:industrialDamageDetect}.

\begin{table*}
\begin{center}
\caption{Examples of DAD techniques used in industrial operations.
        \\CNN: Convolution Neural Networks, LSTM : Long Short Term Memory Networks
        \\GRU: Gated Recurrent Unit, DNN : Deep Neural Networks
        \\AE: Autoencoders, DAE: Denoising Autoencoders, SVM: Support Vector Machines
        \\SDAE: Stacked Denoising Autoencoders, RNN : Recurrent Neural Networks.}
    \label{tab:industrialDamageDetect}
    \captionsetup{justification=centering}
    \scalebox{0.85}{
    \begin{tabular}{ | l | p{2cm} | p{8cm} |}
    \hline
     \textbf{Techniques}  & \textbf{Section} & \textbf{References} \\ \hline
     LSTM & Section ~\ref{sec:rnn_lstm_gru} &  ~\cite{inoue2017anomaly},~\cite{thi2017one},~\cite{kravchik2018detecting},~\cite{huang2018deep},~\cite{park2018lired},~\cite{chang2018review}\\\hline
     AE & Section ~\ref{sec:ae} & ~\cite{yuan2015distributed},~\cite{araya2017ensemble},~\cite{qu2017detection},~\cite{sakurada2014anomaly},~\cite{bhattad2018detecting}\\\hline
     DNN & Section ~\ref{sec:dnn} & ~\cite{lodhi2017power}\\\hline
     CNN & Section ~\ref{sec:cnn} & ~\cite{faghih2016deep},~\cite{christiansen2016deepanomaly},~\cite{lee2016convolutional},~\cite{faghih2016deep}, ~\cite{dong2016camera},~\cite{nanduri2016anomaly},~\cite{fuentes2017robust},~\cite{huang2018deep},~\cite{chang2018review}\\\hline
     SDAE,DAE & Section ~\ref{sec:ae} & ~\cite{yan2015accurate},~\cite{luo2017gas},~\cite{dai2017cleaning} \\\hline
     RNN & Section ~\ref{sec:rnn_lstm_gru} & ~\cite{banjanovic2017neural},~\cite{thi2017one} \\\hline
     Hybrid Models (DNN-SVM) & Section ~\ref{sec:hybridModels} & ~\cite{inoue2017anomaly} \\\hline
    \end{tabular}}
\end{center}
\end{table*}

\label{sec:industrialDamageDetect}


\subsection{Anomaly Detection in Time Series }
\label{sec:timeseriesAD}
Data recorded continuously over duration is known as time series. Time series data can be broadly classified into univariate and multivariate time series. In case of univariate time series, only single variable (or feature) varies over time. For instance, the data collected from a temperature sensor within the room for each second is an uni-variate time series data. A multivariate time series consists several variables (or features) which change over time. An accelerometer which produces three-dimensional data for every second one for each axis $(x,y,z)$ is a perfect example of multivariate time series data. In the literature, types of anomalies in univariate and multivariate time series are categorized into following groups: (1) Point Anomalies.~\ref{sec:PointAnomalies} (2)  Contextual Anomalies~\ref{sec:contextualanomalies} (3) Collective Anomalies ~\ref{sec:groupanomaly}. In recent times, many deep learning models have been proposed for detecting anomalies within univariate and multivariate time series data as illustrated in Table~\ref{tab:univariatesensorAnomalyDetect} and Table ~\ref{tab:multivariatesensorAnomalyDetect} respectively. 
Some of the challenges to detect anomalies in time series using deep learning models data are:
\begin{itemize}
    \item Lack of defined pattern in which an anomaly is occurring may be defined.
    \item Noise within the input data seriously affects the performance of algorithms.
    \item As the length of the time series data increases the computational complexity also increases.
    \item Time series data is usually non-stationary, non-linear and dynamically evolving. Hence DAD models should be able to detect anomalies in real time.
\end{itemize}

\subsubsection{Uni-variate time series deep anomaly detection}
The advancements in deep learning domain offer opportunities to extract rich hierarchical features which can greatly improve outlier detection within uni-variate time series data. The list of industry standard tools and datasets (both deep learning based and non-deep learning based)  for benchmarking anomaly detection algorithms on both univariate and multivariate time-series data is presented and maintained at Github repository~\footnote{\url{https://github.com/rob-med/awesome-TS-anomaly-detection}}.  Table~\ref{tab:univariatesensorAnomalyDetect} illustrates various deep architectures adopted for  anomaly detection within uni-variate time series data.

\begin{table*}
\begin{center}
\caption{Examples of DAD techniques used in uni-variate time series data.
        \\CNN: Convolution Neural Networks, GAN: Generative Adversarial networks,
        \\ DNN: Deep Neural Networks,AE: Autoencoders,DAE: Denoising Autoencoders, 
        \\ VAE: Variational Autoencoder,SDAE: Stacked Denoising Autoencoders, \\ LSTM: Long Short Term Memory Networks, GRU: Gated Recurrent Unit
        \\RNN: Recurrent Neural Networks, RNN: Replicator Neural Networks}
    \captionsetup{justification=centering}
  \label{tab:univariatesensorAnomalyDetect}
  \scalebox{0.90}{
    \begin{tabular}{ | p{3cm} | p{4cm} | p{9cm} |}
    \hline
    \textbf{Techniques}  & \textbf{Section} & \textbf{References} \\ \hline
     LSTM & Section ~\ref{sec:rnn_lstm_gru} & \cite{shipmon2017time},\cite{hundman2018detecting},\cite{zhu2017deep}\cite{malhotra2015long},\cite{chauhan2015anomaly},\cite{assendorp2017deep}\newline \cite{ahmad2017unsupervised},\cite{malhotra2016lstm},\cite{bontemps2016collective},\cite{taylor2016anomaly},\cite{cheng2016ms},\cite{loganathan2018sequence},\cite{chauhan2015anomaly},\cite{malhotra2015long},\cite{gorokhov2017convolutional},~\cite{munir2018deepant}\\\hline
     AE,LSTM-AE,CNN-AE,GRU-AE & Section ~\ref{sec:ae} & ~\cite{Dominique},~\cite{malhotra2016multi},\newline ~\cite{filonov2016multivariate},~\cite{sugimoto2018deep},~\cite{oh2018residual},~\cite{ebrahimzadehmulti},~\cite{veeramachaneni2016ai},~\cite{dau2014anomaly}\\\hline
     RNN & Section ~\ref{sec:rnn_lstm_gru} & ~\cite{wielgosz2017recurrent},~\cite{saurav2018online},~\cite{wielgosz2018model},~\cite{guo2016robust},~\cite{filonov2017rnn}\\\hline
     CNN, CNN-LSTM & Section ~\ref{sec:cnn},~\ref{sec:rnn_lstm_gru} & ~\cite{kanarachos2017detecting},~\cite{dumodeling},~\cite{gorokhov2017convolutional},~\cite{napoletano2018anomaly},~\cite{shanmugam2018jiffy},\cite{medel2016anomaly}\\\hline
     LSTM-VAE & Section ~\ref{sec:rnn_lstm_gru},~\ref{sec:gan_adversarial} & ~\cite{park2018multimodal},~\cite{solch2016variational}\\\hline
     DNN &Section ~\ref{sec:dnn} & ~\cite{amarasinghe2018toward}\\\hline
     GAN &Section ~\ref{sec:gan_adversarial} & ~\cite{li2018anomaly},~\cite{zenati2018efficient},~\cite{lim2018doping},~\cite{laptevanogen},\cite{wei2018unsupervised}\\\hline
    \end{tabular}}
\end{center}
\end{table*}

\subsubsection{Multi-variate time series deep anomaly detection}
Anomaly detection in multivariate time series data is a challenging task. Effective multivariate anomaly detection enables fault isolation diagnostics. RNN and LSTM based methods~\footnote{\url{https://github.com/pnnl/safekit}} are shown to perform well in detecting interpretable anomalies within multivariate time series dataset. DeepAD, a generic framework based on deep learning for multivariate time series anomaly detection is proposed by ~(\cite{buda2018deepad}). Interpretable, anomaly detection systems designed using deep attention based models are effective in explaining the anomalies detected~(\cite{yuan2018muvan,guo2018exploring}). Table ~\ref{tab:multivariatesensorAnomalyDetect} illustrates various deep architectures adopted for  anomaly detection within  multivariate time series data.

\begin{table*}
\begin{center}
\caption{Examples of DAD techniques used in multivariate time series data.
        \\CNN: Convolution Neural Networks, GAN: Generative Adversarial networks,
        \\ DNN: Deep Neural Networks,AE: Autoencoders,DAE: Denoising Autoencoders, 
        \\ VAE: Variational Autoencoder,SDAE: Stacked Denoising Autoencoders, \\ LSTM: Long Short Term Memory Networks, GRU: Gated Recurrent Unit}
    \captionsetup{justification=centering}
  \label{tab:multivariatesensorAnomalyDetect}
  \scalebox{0.90}{
    \begin{tabular}{ | p{3cm} | p{4cm} | p{9cm} |}
    \hline
    \textbf{Techniques}  & \textbf{Section} & \textbf{References} \\ \hline
     LSTM & Section ~\ref{sec:rnn_lstm_gru} & ~\cite{nucci2018real},~\cite{hundman2018detecting},~\cite{Assendorp2017DeepLF},~\cite{nolle2018binet}\\\hline
     AE,LSTM-AE,CNN-AE,GRU-AE & Section ~\ref{sec:ae} & ~\cite{zhang2018deep}~\cite{guo2018multidimensional},~\cite{fu2019aircraft},~\cite{kieu2018outlier}\\\hline
     CNN, CNN-LSTM & Section ~\ref{sec:cnn},~\ref{sec:rnn_lstm_gru} & ~\cite{basumallik2019packet},~\cite{shanmugam2018jiffy}\\\hline
     LSTM-VAE & Section ~\ref{sec:rnn_lstm_gru},~\ref{sec:gan_adversarial} & ~\cite{ikeda2018estimation},~\cite{park2018multimodal}\\\hline
     GAN &Section ~\ref{sec:gan_adversarial} & ~\cite{assendorp2017deep},~\cite{li2018anomaly},~\cite{li2019mad}~\cite{cowton2018combined}\\\hline
     DNN-RNN & Section ~\ref{sec:rnn_lstm_gru}& ~\cite{tuor2017deep},~\cite{tuor2018recurrent}\\\hline
    \end{tabular}}
\end{center}
\end{table*}



\begin{table}[!htbp]
 \centering
\caption{Datasets used in multivariate anomaly detection.}
  \begin{threeparttable}[t]
  \centering
    \begin{tabular}{ | p{3cm} | p{5cm} | p{5cm} |}
    \toprule
     Dataset     & Description  &   References \\
    \midrule
    NASA Shuttle Valve Data\tnote{1}   & Includes spacecraft anomaly data  and experiments from the Mars Science Laboratory and SMAP missions  & \cite{hundman2018detecting}\tnote{2} \\ \hline 
    Vessels\tnote{3}  & Multivariate temporal data analysis for
    Vessels behavior anomaly detection    & \cite{multivariate17} \\
    \hline 
    SWaT and WADI &
     Secure Water Treatment (SWaT) and the Water Distribution (WADI)   & \cite{li2019mad} \\
    \hline 
     Credit Card Fraud Detection &
     Anonymized credit card transactions labeled as fraudulent or genuine   & \cite{dal2015calibrating} \\
    \hline 
    NYC taxi passenger count\tnote{5} &
     The New York City taxi passenger data stream   & \cite{cui2016comparative} \\
    \hline 
    
     \bottomrule
  \end{tabular}
     \begin{tablenotes}
     \item[1]  \url{https://cs.fit.edu/~pkc/nasa/data/}
     \item[2]  \url{https://github.com/khundman/telemanom}
     \item[3] \url{http://conferences.inf.ed.ac.uk/EuroDW2018/papers/eurodw18-Maia.pdf}
     \item[4] \url {https://www.kaggle.com/peterkim95/multivariate-gaussian-anomaly-detection/data}
     \item[5] \url{https://github.com/chickenbestlover/RNN-Time-series-Anomaly-Detection}
   \end{tablenotes}
    \end{threeparttable}%
  \label{tab:addlabel}%
\end{table}%


\label{sec:sensorNetworkAnomaly}


\begin{table*}
\begin{center}
\caption{Examples of DAD techniques used in video surveillance.
        \\CNN: Convolution Neural Networks, LSTM : Long Short Term Memory Networks
        \\RBM: Restricted Boltzmann Machine, DNN : Deep Neural Networks
        \\AE: Autoencoders, DAE: Denoising Autoencoders
        \\OCSVM: One class Support vector machines, CAE: Convolutional Autoencoders
        \\SDAE: Stacked Denoising Autoencoders, STN : Spatial Transformer Networks }
  \label{tab:videoSurvellianceAnomalyDetect}
  \captionsetup{justification=centering}
  \scalebox{0.80}{
    \begin{tabular}{ | p{3cm} | p{4cm} | p{12cm} |}
      \hline
      \textbf{Technique Used} & \textbf{Section} & \textbf{References}\\
      \hline
      CNN & Section ~\ref{sec:cnn} & \cite{dong2016camera},\cite{andrewsaanomaly},\cite{sabokrou2016fully},\cite{sabokrou2017deep},\cite{munawar2017spatio},\cite{li2017transferred},\cite{qiao2017abnormal},\cite{tripathi2018convolutional},\cite{nogas2018deepfall},\cite{christiansen2016deepanomaly},\cite{li2017transferred},\cite{}\\\hline
      SAE (AE-CNN-LSTM)  &  Section ~\ref{sec:ae},~\ref{sec:cnn},~\ref{sec:rnn_lstm_gru}  & ~\cite{chong2017abnormal},~\cite{qiao2017abnormal},~\cite{khaleghi2018improved}\\\hline
      AE &  Section ~\ref{sec:ae}  & \cite{qiao2017abnormal},\cite{yang2015unsupervised},\cite{chen2015detecting},\cite{gutoskidetection},\cite{d2017autoencoder},\cite{dotti2017unsupervised},\cite{yang2015unsupervised},\cite{chen2015detecting},\cite{sabokrou2016video},\cite{tran2017anomaly},\cite{chen2015detecting} ,\cite{d2017autoencoder},\cite{hasan2016learning},\cite{yang2015unsupervised},\cite{cinelli2017anomaly},\cite{sultani2018real}\\\hline
      Hybrid Model (CAE-OCSVM) & Section ~\ref{sec:hybridModels}  & ~\cite{gutoskidetection}, ~\cite{dotti2017unsupervised}\\\hline
      LSTM-AE &  Section ~\ref{sec:rnn_lstm_gru}, ~\ref{sec:ae}  & ~\cite{d2017autoencoder}\\\hline
      STN &Section~\ref{sec:stn}   & \cite{chianucci2016unsupervised}\\\hline
      RBM &Section ~\ref{sec:dnn}   & \cite{munawar2017spatio}\\\hline
      LSTM &Section ~\ref{sec:rnn_lstm_gru}  &~\cite{medel2016anomaly},~\cite{luo2017remembering},~\cite{ben2018attentioned},~\cite{singh2017anomaly}\\\hline
      RNN & Section ~\ref{sec:rnn_lstm_gru} &\cite{luo2017revisit},\cite{zhou2015abnormal} ,\cite{hu2016video},~\cite{chong2015modeling}\\\hline
      AAE & Section ~\ref{sec:gan_adversarial} & ~\cite{ravanbakhsh2017training}\\\hline
    \end{tabular}}
  \end{center}
\end{table*}

\subsection{Video Surveillance}
Video Surveillance also popularly known as Closed-circuit television (CCTV) involves monitoring designated areas of interest in order to ensure security. In videos surveillance applications unlabelled data is available in large amounts, this is a significant challenge for supervised machine learning and deep learning methods. Hence video surveillance applications have been modeled as anomaly detection problems owing to lack of availability of labeled data. Several works have studied the state-of-the-art deep models for video anomaly detection and have classified them based on the type of model and criteria of detection ~(\cite{kiran2018overview,chong2015modeling}). The challenges of robust 24/7 video surveillance systems are discussed in detail by  ~(\cite{boghossian2005challenges}). The lack of an explicit definition of an anomaly in real-life video surveillance is a significant issue that hampers the performance of DAD methods as well. DAD techniques used in video surveillance are illustrated in Table ~\ref{tab:videoSurvellianceAnomalyDetect}.


\label{sec:videoSurvelliance}

\section{Deep Anomaly Detection (DAD) Models}
\label{sec:deepDADModels}

In this section, we discuss various DAD models classified based on the availability of labels and training objective. For each model types domain, we discuss the following four aspects:\\
\textemdash assumptions;\\
\textemdash type of model architectures;\\
\textemdash computational complexity;\\
\textemdash advantages and disadvantages;\\

\subsection{Supervised deep anomaly detection}
\label{sec:supervisedDAD}
 Supervised anomaly detection techniques are superior in performance compared to unsupervised anomaly detection techniques  since these techniques use  labeled samples~(\cite{gornitz2013toward}).  Supervised anomaly detection learns the separating boundary from a set of annotated data instances (training) and then, classify a test instance into  either normal or anomalous classes with the learned model (testing).\\
\textbf{Assumptions:}
Deep supervised learning methods depend on separating data classes whereas unsupervised
techniques focus on explaining and understanding the characteristics of data. Multi-class classification based anomaly detection techniques assumes that the training data contains labeled instances of  multiple normal classes ~(\cite{shilton2013combined,jumutc2014multi,kim2015deep,erfani2017shared}). Multi-class anomaly detection techniques learn a classifier to distinguish between anomalous class from the rest of the classes. In general, supervised deep learning-based classification schemes for anomaly detection have two sub-networks, a feature extraction network followed by a classifier network. Deep models require a substantial number of training samples (in the order of thousands or millions) to learn feature representations to discriminate various class instances effectively. Due to, lack of availability of clean data labels supervised deep anomaly detection techniques are not so popular as semi-supervised and unsupervised methods.\\
\textbf{Computational Complexity:} 
The computational complexity of deep supervised anomaly detection methods based techniques depends on the input data dimension and the number of hidden layers trained using back-propagation algorithm. High dimensional data tend to have more hidden layers to ensure meaning-full hierarchical learning of input features.The computational complexity also increases linearly with the number of hidden layers and require greater model training and update time.

\textbf{Advantages and Disadvantages:}
The advantages of supervised DAD techniques are as follows:
\begin{itemize}
\item Supervised DAD methods are more accurate than semi-supervised and unsupervised models.
\item The testing phase of classification based techniques is fast since each test instance needs to be compared against the precomputed model.
\end{itemize}
The disadvantages of Supervised DAD techniques are as follows:
\begin{itemize}
\item  Multi-class supervised techniques require accurate labels for various normal classes and anomalous instances, which is often not available.
\item Deep supervised techniques fail to separate normal from anomalous data if the feature space is highly complex and non-linear.
\end{itemize}

\label{sec:supervised}

\subsection{Semi-supervised deep anomaly detection }
\label{sec:semi_supervised_DAD}
Semi-supervised or (one-class classification) DAD techniques assume that all training instances have only one class label.  A review of deep learning based semi-supervised techniques for anomaly detection is presented by ~\cite{kiran2018overview} and \cite{min2018ids}. DAD techniques learn a discriminative boundary around the normal instances. The test instance that does not belong to the majority class is flagged as being anomalous~(\cite{perera2018learning,blanchard2010semi}). Various  semi-supervised DAD model architectures are illustrated in Table~\ref{tab:semisupervisedModels}.

\begin{table*}
\begin{center}
\caption{Semi-supervised DAD models overview
        \\AE: Autoencoders, DAE: Denoising Autoencoders, KNN : K- Nearest Neighbours
        \\CorGAN: Corrupted Generative Adversarial Networks, DBN: Deep Belief Networks
        \\ AAE: Adversarial Autoencoders, CNN: Convolution neural networks
        \\ SVM:  Support vector machines.}
    \label{tab:semisupervisedModels}
    \begin{tabular}{ | p{4cm} | p{4cm} | p{4cm} |}
    \hline
     \textbf{Techniques}  & \textbf{Section} & \textbf{References} \\ \hline
     AE & Section ~\ref{sec:ae} & ~\cite{edmunds2017deep} ,~\cite{estiri2018semi}\\\hline
     RBM & Section ~\ref{sec:dnn} & ~\cite{jia2014novel} \\\hline
     DBN & Section ~\ref{sec:dnn} & ~\cite{wulsin2010semi},~\cite{wulsin2011modeling} \\\hline
     CorGAN,GAN & Section~\ref{sec:gan_adversarial} & ~\cite{gu2018semi} ~\cite{akcay2018ganomaly},~\cite{sabokrou2018adversarially}\\\hline
     AAE &Section~\ref{sec:gan_adversarial} & ~\cite{dimokranitou2017adversarial}\\\hline
     Hybrid Models (DAE-KNN~\cite{altman1992introduction}), (DBN-Random Forest~\cite{ho1995random}),CNN-Relief~\cite{kira1992feature},CNN-SVM~\cite{cortes1995support} & Section~\ref{sec:DHM} & ~\cite{song2017hybrid},~\cite{shi2017semi},~\cite{zhu2018hybrid} \\\hline
     CNN & Section~\ref{sec:cnn} & ~\cite{racah2017extremeweather},~\cite{perera2018learning} \\ \hline
     RNN & Section~\ref{sec:rnn_lstm_gru} & ~\cite{wu2018semi} \\ \hline
     GAN & Section~\ref{sec:gan_adversarial} & ~\cite{kliger2018novelty},~\cite{gu2018semi} \\ \hline
    \end{tabular}
\end{center}
\end{table*}


\textbf{Assumptions: }
 Semi-supervised DAD methods proposed to rely on one of the following assumptions to score a data instance as an anomaly.
\begin{itemize}
 \item Proximity and Continuity: Points which are close to each other both in input space and learned feature space are more likely to share the same label.
  \item Robust features are learned within hidden layers of deep neural network layers and retain the discriminative attributes for separating normal from outlier data points.
\end{itemize}

\textbf{Computational Complexity:} 
The computational complexity of semi-supervised DAD methods based techniques is similar to supervised DAD techniques, which primarily depends on the dimensionality of the input data and the number of hidden layers used for representative feature learning.\\

\textbf{Advantages and Disadvantages:}
The advantages of semi-supervised deep anomaly detection techniques are as follows:
\begin{itemize}
\item  Generative Adversarial Networks (GANs) trained in semi-supervised learning mode have shown great promise, even with very few labeled data.
\item  Use of labeled data ( usually of one class), can produce considerable performance improvement   over unsupervised techniques.
\end{itemize}
The fundamental disadvantages of semi-supervised techniques presented by~(\cite{lu2009fundamental}) are applicable even in a deep learning context. Furthermore, the hierarchical features extracted within hidden layers may not be representative of fewer anomalous instances hence are prone to the over-fitting problem.

\label{sec:semiSupervised}


\begin{table*}
\begin{center}
\caption{Examples of  Hybrid DAD techniques.
        \\CNN: Convolution Neural Networks, LSTM : Long Short Term Memory Networks
        \\DBN: Deep Belief Networks, DNN : Deep Neural Networks.
        \\AE: Autoencoders, DAE: Denoising Autoencoders, SVM: Support Vector Machines~\cite{cortes1995support}
        \\SVDD: Support Vector Data Description, RNN : Recurrent Neural Networks
        \\Relief: Feature selection Algorithm~\cite{kira1992feature}, KNN: K- Nearest Neighbours~\cite{altman1992introduction}
        \\CSI: Capture, Score, and Integrate~\cite{ruchansky2017csi}. }
    \captionsetup{justification=centering}
    \label{tab:hybridModels}
    \scalebox{0.85}{
    \begin{tabular}{ | p{3cm} | p{2cm} | p{6cm} |}
    \hline
     \textbf{Techniques}  & \textbf{Section} & \textbf{References} \\ \hline
     AE-OCSVM, AE-SVM & Section ~\ref{sec:ae}, & ~\cite{andrews2016detecting} \\\hline
     DBN-SVDD, AE-SVDD & Section ~\ref{sec:dnn}, & ~\cite{erfani2016high},~\cite{kim2015deep} \\\hline
     DNN-SVM & 21D & ~\cite{inoue2017anomaly} \\\hline
     DAE-KNN, DBN-Random Forest~\cite{ho1995random},CNN-Relief,CNN-SVM & Section ~\ref{sec:dnn},\ref{sec:ae} & ~\cite{song2017hybrid},~\cite{shi2017semi},~\cite{zhu2018hybrid,urbanowicz2018relief} \\\hline
     AE-CNN, AE-DBN & Section ~\ref{sec:dnn},~\ref{sec:cnn},\ref{sec:ae} &  ~\cite{wang2018effective},~\cite{li2015hybrid} \\\hline
     AE+ KNN & Section \ref{sec:ae} & ~\cite{song2017hybrid} \\\hline
     CNN-LSTM-SVM & Section ~\ref{sec:cnn},\ref{sec:rnn_lstm_gru}  & ~\cite{wei2017new}\\
     RNN-CSI & Section ~\ref{sec:rnn_lstm_gru} & ~\cite{ruchansky2017csi}\\
     CAE-OCSVM & Section \ref{sec:ae} & ~\cite{gutoskidetection}, ~\cite{dotti2017unsupervised}\\\hline
    \end{tabular}}
\end{center}
\end{table*}

\subsection{Hybrid deep anomaly detection}
\label{sec:hybridModels}

Deep learning models are widely used as feature extractors to learn robust features ~(\cite{andrews2016detecting}). In deep hybrid models, the representative features learned within deep models are input to traditional algorithms like one-class Radial Basis Function (RBF), Support Vector Machine (SVM) classifiers. The hybrid models employ two step learning and are shown to produce state-of-the-art results ~(\cite{erfani2016high,erfani2016robust,wu2015harvesting}). Deep hybrid architectures used in anomaly detection is presented in  Table ~\ref{tab:hybridModels}.



\textbf{Assumptions: } \\
The deep hybrid models proposed for anomaly detection rely on one of the following assumptions to detect outliers:
\begin{itemize}
  \item Robust features are extracted within hidden layers of the deep neural network, aid in separating the irrelevant features which can conceal the presence of anomalies.
  \item Building a robust anomaly detection model on complex, high-dimensional spaces require feature extractor and an anomaly detector. Various anomaly detectors used alongwith are illustrated in Table ~\ref{tab:hybridModels}
\end{itemize}

\textbf{Computational Complexity :} \\
The computational complexity of a hybrid model includes the complexity of both deep architectures as well as traditional algorithms used within. Additionally, an inherent issue of non-trivial choice of deep network architecture and parameters which involves searching optimized parameters in a considerably larger space introduces the computational complexity of using deep layers within hybrid models. Furthermore considering the classical algorithms such as linear SVM which has prediction complexity  of $O(d)$ with d the number of input dimensions. For most kernels, including polynomial and RBF, the complexity is $O(nd)$ where $n$ is the number of support vectors although an approximation $O(d^2)$ is considered for SVMs with an RBF kernel.

\textbf{Advantages and Disadvantages }\\
The advantages of hybrid DAD techniques are as follows:
\begin{itemize}
\item  The feature extractor significantly reduces the ‘curse of dimensionality', especially in the high dimensional domain.
\item  Hybrid models are more scalable and computationally efficient since the linear or nonlinear kernel models operate on reduced input dimension.
\end{itemize}
The significant disadvantages of hybrid DAD techniques are:
\begin{itemize}
\item  The hybrid approach is suboptimal because it is unable to influence representational learning within the hidden layers of feature extractor since generic loss functions are employed instead of the customized objective for anomaly detection.
\item The deeper hybrid models tend to perform better if the individual layers are ~(\cite{saxe2011random}) which introduces computational expenditure.
\end{itemize}

\label{sec:hybrid}

\subsection{One-class neural networks (OC-NN) for anomaly detection}
\label{sec:oneclassNN}
One-class neural networks (OC-NN) combines the ability of deep networks to extract a progressively rich representation of data alongwith the one-class objective, such as a hyperplane~(\cite{chalapathy2018anomaly})  or hypersphere ~(\cite{ruff2018deep}) to separate all the normal data points from the outliers. The OC-NN approach is novel for the following crucial reason: data representation in the hidden layer are learned by optimizing the objective function customized for anomaly detection as illustrated in The experimental results in ~(\cite{chalapathy2018anomaly,ruff2018deep}) demonstrate that OC-NN can achieve comparable or better performance than existing state-of-the-art methods for complex datasets, while having reasonable training and testing time compared to the existing methods.

\textbf{Assumptions:} 
The OC-NN models proposed for anomaly detection rely on the  following assumptions to detect outliers:
\begin{itemize}
 \item  OC-NN models extract the common factors of variation within the data distribution within the hidden layers of the deep neural network.
  \item Performs combined representation learning and produces an outlier score for a test data instance.
  \item Anomalous samples do not contain common factors of variation and hence hidden layers fail to capture the representations of outliers.
\end{itemize}

\textbf{Computational Complexity:} 
The Computational complexity of an OC-NN model as against the hybrid model includes only the complexity of the deep network of choice ~(\cite{saxe2011random}). OC-NN models do not require data to be stored for prediction, thus have very low memory complexity. However, it is evident that the OC-NN training time is proportional to the input dimension.

\textbf{Advantages and Disadvantages:}
The advantages of OC-NN  are as follows:
\begin{itemize}
\item  OC-NN models jointly train a deep neural network while optimizing a data-enclosing hypersphere or hyper-plane in output space.
\item OC-NN propose an alternating minimization algorithm for learning the parameters of the OC-NN model. We observe that the subproblem of the OC-NN objective is equivalent to a solving a quantile selection problem which is well defined.
\end{itemize}
The significant disadvantages of OC-NN for anomaly detection are:
\begin{itemize}
\item Training times  and model update time may be longer for high dimensional input data.
\item Model updates would also take longer time, given the change in input space.
\end{itemize}

\label{sec:oneClassNeuralNetworks}

\subsection{Unsupervised Deep Anomaly Detection }
\label{sec:unsupervisedDAD}
Unsupervised DAD is an essential area of research in both fundamental machine learning research and industrial applications. Several deep learning frameworks that address challenges in unsupervised anomaly detection are proposed and shown to produce a state-of-the-art performance as illustrated in Table ~\ref{tab:unsupervisedAnomalyDetection}. Autoencoders are the fundamental unsupervised deep architectures used in anomaly detection ~(\cite{baldi2012autoencoders}).

\begin{table*}
\begin{center}
\caption{Examples of  Un-supervised DAD techniques .
        \\CNN: Convolution Neural Networks, LSTM : Long Short Term Memory Networks
        \\DNN : Deep Neural Networks., GAN: Generative Adversarial Network
        \\AE: Autoencoders, DAE: Denoising Autoencoders, SVM: Support Vector Machines
        \\STN: Spatial Transformer Networks, RNN : Recurrent Neural Networks
        \\AAE: Adversarial Autoencoders, VAE : Variational Autoencoders.}
\captionsetup{justification=centering}
    \label{tab:unsupervisedAnomalyDetection}
    \scalebox{0.85}{
    \begin{tabular}{ | p{3cm}  | p{2cm} | p{8cm} |}
    \hline
     \textbf{Techniques}  & \textbf{Section} & \textbf{References} \\ \hline
     LSTM & Section ~\ref{sec:rnn_lstm_gru} &  ~\cite{singh2017anomaly},~\cite{chandola2008comparative},~\cite{dasigi2014modeling},\cite{malhotra2015long}\\\hline
     AE & Section ~\ref{sec:ae} & ~\cite{abati2018and},~\cite{zong2018deep},~\cite{tagawa2015structured},~\cite{dau2014anomaly},~\cite{sakurada2014anomaly},~\cite{wu2015adaptive},\newline~\cite{xu2015learning},~\cite{hawkins2002outlier},~\cite{zhao2015robust},~\cite{qi2014robust},~\cite{chalapathy2017robust},~\cite{yang2015unsupervised},\newline ~\cite{zhai2016deep},~\cite{lyudchik2016outlier},~\cite{lu2017unsupervised},~\cite{mehrotra2017deep},~\cite{meng2018relational},\cite{parchami2017using}\\\hline
     STN & Section ~\ref{sec:stn} & ~\cite{chianucci2016unsupervised}\\\hline
     GAN & Section ~\ref{sec:gan_adversarial} & ~\cite{lawson2017finding} \\\hline
     RNN & Section ~\ref{sec:rnn_lstm_gru} & ~\cite{dasigi2014modeling},\cite{filonov2017rnn} \\\hline
     AAE & Section ~\ref{sec:gan_adversarial} & ~\cite{dimokranitou2017adversarial},~\cite{leveau2017adversarial} \\\hline
     VAE & Section ~\ref{sec:gan_adversarial} &  ~\cite{an2015variational},~\cite{suh2016echo},~\cite{solch2016variational},~\cite{xu2018unsupervised},~\cite{mishra2017generative}\\\hline
    \end{tabular}}
\end{center}
\end{table*}


\textbf{Assumptions: } 
The deep unsupervised models proposed for anomaly detection rely on one of the following assumptions to detect outliers:
\begin{itemize}
 \item The “normal” regions in the original or  latent feature space can be distinguished from "anomalous" regions in the original or  latent feature space.
  \item The majority of the data instances are normal compared to the remainder of the data set.
  \item Unsupervised anomaly detection algorithm produces an outlier score of the data instances based on  intrinsic properties of the data-set such as distances or densities. The hidden layers of deep neural network aim to capture these intrinsic properties within the dataset~(\cite{goldstein2016comparative}).
\end{itemize}

\textbf{Computational Complexity:} 
The autoencoders are the most common architecture employed in outlier detection with quadratic cost, the optimization problem is non-convex, similar to any other neural network architecture. The  computational complexity of model depends on the number of operations, network parameters, and hidden layers. However, the computational complexity of training an autoencoder is much higher than traditional methods such as Principal Component Analysis (PCA) since PCA is based on matrix decomposition ~(\cite{meng2018relational,parchami2017using}).\\

\textbf{Advantages and Disadvantages:}
The advantages of unsupervised deep anomaly detection techniques are as follows:
\begin{itemize}
\item  Learns the inherent data characteristics to separate normal from an anomalous data point. This technique identifies commonalities within the data and facilitates outlier detection.
\item  Cost effective technique to find the anomalies since it does not require annotated data for training the algorithms.
\end{itemize}
The significant disadvantages of unsupervised deep anomaly detection techniques are:
\begin{itemize}
\item  Often it is challenging to learn commonalities within data in a complex and high dimensional space.
\item While using autoencoders the choice of right degree of compression, i.e., dimensionality reduction is often an hyper-parameter that requires tuning for optimal results.
\item Unsupervised techniques techniques are very sensitive to noise, and data corruptions and are often less accu-rate than supervised or semi-supervised techniques.
\end{itemize}

\label{sec:unsupervised}

\subsection{Miscellaneous Techniques}
This section explores, various DAD techniques which are shown to be effective and promising,  we discuss the key idea behind those techniques and their area of applicability.

\subsubsection{Transfer Learning based anomaly detection }
Deep learning for long has been criticized for the need to have enough data to produce good results. Both \cite{litjens2017survey} and \cite{pan2010survey} present the review of deep transfer learning approaches and illustrate their significance to learn good feature representations. Transfer learning is an essential tool in machine learning to solve the fundamental problem of insufficient training data. It aims to transfer the knowledge from the source domain to the target domain by relaxing the assumption that training and future data must be in the same feature space and have the same distribution. Deep transfer representation-learning has been explored by ~(\cite{andrews2016transfer,vercruyssen2017transfer,li2012detecting,almajai2012anomaly,kumar2017transfer,liang2018transfer}) are shown to produce very promising results. The open research questions using transfer learning for anomaly detection is, the degree of transfer-ability, that is to define how well features transfer the knowledge and improve the classification performance from one task to another.

\subsubsection{Zero Shot learning based anomaly detection}
Zero shot learning (ZSL) aims to recognize objects never seen before within training set ~(\cite{romera2015embarrassingly}). ZSL achieves this in two phases: Firstly the knowledge about the objects in natural language descriptions or attributes (commonly known as meta-data) is captured Secondly this knowledge is then used to classify instances among a new set of classes. This setting is important in the real world since one may not be able to obtain images of all the possible classes at training. The primary challenge associated with this approach is the obtaining the meta-data about the data instances. However several approaches of using ZSL in anomaly and novelty detection are shown to produce state-of-the-art results ~(\cite{mishra2017generative,socher2013zero,xian2017zero,liu2017generalized,rivero2017grassmannian}).

\subsubsection{Ensemble based anomaly detection}
A notable issue with deep neural networks is that they are sensitive to noise within input data and often require extensive training data to perform robustly~(\cite{kim2016lstm}). In order to achieve robustness even in noisy data an idea to randomly vary on the connectivity architecture of the autoencoder is shown to obtain significantly better performance. Autoencoder ensembles consisting of various randomly connected autoencoders are experimented by ~\cite{chen2017outlier} to achieve promising results on several benchmark datasets. The ensemble approaches are still an active area of research which has been shown to produce improved diversity, thus avoid overfitting problem while reducing training time.

\subsubsection{Clustering based anomaly detection}
Several anomaly detection  algorithms based on clustering have been proposed in literature (\cite{ester1996density}). Clustering involves grouping together similar patterns based on features extracted  detect new anomalies. The time and space complexity grows linearly with number of classes to be clustered ~(\cite{sreekanth2010generalized}), which renders the clustering based anomaly detection prohibitive for real-time practical applications.  The dimensionality of the input data is reduced  extracting features within the hidden layers of deep neural network which ensures scalability for complex and high dimensional datasets.  Deep learning enabled clustering approach anomaly detection utilizes e.g word2vec ~(\cite{mikolov2013efficient})  models to get the semantical presentations of normal data and anomalies to form clusters and detect outliers ~(\cite{yuan2017deep}). Several works rely on variants of hybrid models along with auto-encoders for obtaining representative features for  clustering to find anomalies.

\subsubsection{Deep Reinforcement Learning (DRL) based anomaly detection}
\label{reinforcementlearning}
Deep reinforcement learning (DRL) methods have attracted significant interest due to its ability to learn complex behaviors in high-dimensional data space. Efforts to detect anomalies using deep reinforcement learning have been proposed by ~\cite{de2017learning,rlanomaly}.
The DRL based anomaly detector  does not consider any assumption about the concept of the anomaly,  the detector identifies new anomalies by consistently enhancing its knowledge  through reward signals accumulated. DRL based anomaly detection is a very novel concept which  requires further investigation and identification of the research gap and its applications.

\subsubsection{Statistical techniques deep anomaly detection }

Hilbert transform is a statistical signal processing technique which derives the analytic representation of a real-valued signal. This property is leveraged by ~(\cite{kanarachos2015anomaly}) for real-time detection of anomalies in health-related time series dataset and is shown to be a very promising technique. The algorithm combines the ability of wavelet analysis, neural networks and Hilbert transform in a sequential manner to detect real-time anomalies. The topic of statistical techniques DAD techniques requires further investigation to understand their potential and applicability for anomaly detections fully.

\label{sec:others}

\section{Deep neural network architectures for locating anomalies}
\label{sec:locatingAnomalieswithNNArchitecture}

\subsection{Deep Neural Networks (DNN)}
\label{sec:dnn}

The "deep" in "deep neural networks" refers to the number of layers through which the features of data are extracted ~(\cite{schmidhuber2015deep,bengio2009learning}). Deep architectures overcome the limitations of traditional machine learning approaches of scalability, and generalization to new variations within data~(\cite{lecun2015deep}) and the need for manual feature engineering. Deep Belief Networks (DBNs) are class of deep neural network which comprises multiple layers of graphical models known as Restricted Boltzmann Machine (RBMs). The hypothesis in using DBNs for anomaly detection is that RBMs are used as a directed encoder-decoder network with backpropagation algorithm ~(\cite{werbos1990backpropagation}). DBNs fail to capture the characteristic variations of anomalous samples, resulting in high reconstruction error. DBNs are shown to scale efficiently to big-data and improve interpretability ~(\cite{wulsin2010semi}).

\subsection{Spatio Temporal Networks (STN)}
\label{sec:stn}

Researchers for long have explored techniques to learn both spatial and temporal relation features ~(\cite{zhang2018detecting}). Deep learning architectures is leveraged to perform well at learning spatial aspects ( using CNN's) and temporal features ( using LSTMs) individually. Spatio Temporal Networks (STNs) comprises of deep neural architectures combining both CNN's and LSTMs to extract spatiotemporal features. The temporal features (modeling correlations between near time points via LSTM), spatial features (modeling local spatial correlation via local CNN's) are shown to be effective in detecting outliers ~(\cite{lee2018stan,szeker2014spatio,nie2018spatio,dereszynski2011spatiotemporal}).

\subsection{Sum-Product Networks (SPN)}
\label{sec:spn}
Sum-Product Networks (SPNs) are directed acyclic graphs with variables as leaves, and the internal nodes, and weighted edges constitute the sums and products. SPNs are considered as a combination of mixture models  which have fast exact probabilistic inference over many layers~(\cite{poon2011sum,peharz2018probabilistic}). The main advantage of SPNs is that, unlike graphical models, SPNs are more traceable over high treewidth models without requiring approximate inference. Furthermore, SPNs are shown to capture uncertainty over their inputs in a convincing manner, yielding robust anomaly detection~(\cite{peharz2018probabilistic}). SPNs are shown to be  impressive results on numerous datasets, while much remains to be further explored in relation to outlier detection.

\subsection{Word2vec Models}
\label{sec:word2vec}

Word2vec is a group of deep neural network models used to produce word embeddings ~(\cite{mikolov2013efficient}). These models are capable of capturing sequential relationships within data instance such as sentences, time sequence data. Obtaining word embedding features as inputs are shown to improve the performance in several deep learning architectures ~(\cite{rezaeinia2017improving,naili2017comparative,altszyler2016comparative}). Anomaly detection models leveraging the word2vec embeddings are shown to significantly improve performance~(\cite{schnabel2015evaluation,bertero2017experience,bakarov2018anomaly,bamler2017dynamic}).

\subsection{Generative Models }
\label{sec:gan_adversarial}

Generative models aim to learn exact data distribution in order to generate new data points with some variations. The two most common and efficient generative approaches are Variational Autoencoders (VAE) ~(\cite{kingma2013auto}) and Generative Adversarial Networks (GAN)~(\cite{NIPS2014_5423,goodfellow2014generative}). A variant of GAN architecture known as Adversarial autoencoders (AAE) (~\cite{makhzani2015adversarial}) that use adversarial training to impose an arbitrary prior on the latent code learned within hidden layers of autoencoder are also shown to learn the input distribution effectively. Leveraging this ability of learning input distributions, several Generative Adversarial Networks-based Anomaly Detection (GAN-AD) frameworks ~(\cite{li2018anomaly,deecke2018anomaly,schlegl2017unsupervised,ravanbakhsh2017abnormal,eide2018applying}) proposed are shown to be effective in identifying anomalies on high dimensional and complex datasets. However traditional methods such as K-nearest neighbors (KNN) are shown to perform better in scenarios which have a lesser number of anomalies when compared to deep generative models ~(\cite{vskvara2018generative}).

\subsection{Convolutional Neural Networks }
\label{sec:cnn}
Convolutional Neural Networks (CNN), are the popular choice of neural networks for analyzing visual imagery ~(\cite{krizhevsky2012imagenet}). CNN's ability to extract complex hidden features from high dimensional data with complex structure has enabled its use as feature extractors in outlier detection for both sequential and image dataset ~(\cite{gorokhov2017convolutional,kim2014convolutional}). Evaluation of CNN's based frameworks for anomaly detection is currently still an active area of research ~(\cite{kwon2018empirical}).

\subsection{Sequence Models}
\label{sec:rnn_lstm_gru}

Recurrent Neural Networks (RNNs) ~(\cite{williams1989complexity}) are shown to capture features of time sequence data. The limitations with RNNs is that they fail to capture the context as time steps increases, in order to resolve this problem, Long Short-Term Memory ~(\cite{hochreiter1997long}) networks were introduced, they are a particular type of RNNs comprising of a memory cell that can store information about previous time steps. Gated Recurrent Unit ~(\cite{cho2014learning}) (GRU) are similar to LSTMs, but use a set of gates to control the flow of information, instead of separate memory cells. Anomaly detection in sequential data has attracted significant interest in the literature due to its applications in a wide range of engineering problems illustrated in Section ~\ref{sec:timeseriesAD}. Long Short Term Memory (LSTM) neural network based algorithms for anomaly detection have been investigated and reported to produce significant performance gains over conventional methods ~(\cite{ergen2017unsupervised}).

\subsection{Autoencoders}
\label{sec:ae}

Autoencoders with single layer along with a linear activation function are nearly equivalent to Principal Component Analysis (PCA)~(\cite{pearson1901liii}). While PCA is restricted to a linear dimensionality reduction, auto encoders enable both linear or nonlinear tranformations~(\cite{liou2008modeling,liou2014autoencoder}). One of the popular applications of Autoencoders is anomaly detection. Autoencoders are also referenced by the name Replicator Neural Networks (RNN) ~(\cite{hawkins2002outlier},~\cite{williams2002comparative}). Autoencoders represent data within multiple hidden layers by reconstructing the input data, effectively learning an identity function. The autoencoders, when trained solely on normal data instances ( which are the majority in anomaly detection tasks), fail to reconstruct the anomalous data samples, therefore, producing a large reconstruction error. The data samples which produce high residual errors are considered outliers. Several variants of autoencoder architectures are proposed as illustrated in Figure ~\ref{fig:aevariants} produce promising results in anomaly detection. The choice of autoencoder architecture depends on the nature of data, convolution networks are preferred for image datasets while Long short-term memory (LSTM) based models tend to produce good results for sequential data. Efforts to combine both convolution and LSTM layers where the encoder is a convolutional neural network (CNN) and decoder is a multilayer LSTM network to reconstruct input images are shown to be effective in detecting anomalies within data. The use of combined models such as Gated recurrent unit autoencoders (GRU-AE), Convolutional neural networks autoencoders (CNN-AE), Long short-term memory (LSTM) autoencoder (LSTM-AE) eliminates the need for preparing hand-crafted features and facilitates the use of raw data with minimal preprocessing in anomaly detection tasks. Although autoencoders are simple and effective architectures for outlier detection, the performance gets degraded  due to noisy training data ~(\cite{zhou2017anomaly}).

\tikzset{edge from parent/.style=
{draw, edge from parent path={(\tikzparentnode.south)
-- +(0,-8pt)
-| (\tikzchildnode)}},
blank/.style={draw=none}}

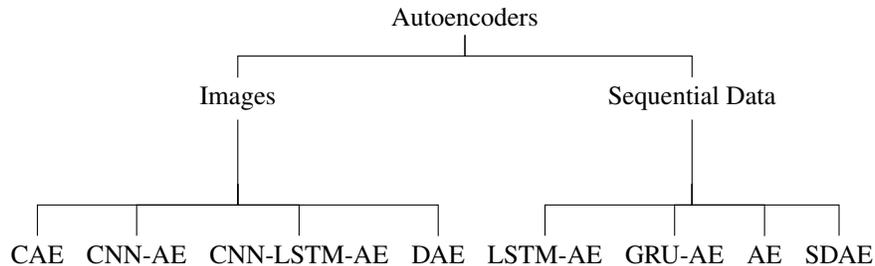
\begin{figure}
\centering
\begin{tikzpicture}
\matrix
{
&
\node{\Tree
 [.Autoencoders
    [.Images
        [{CAE} {CNN-AE}  {CNN-LSTM-AE} {DAE} ] ]
    [.{Sequential Data}
            [ {LSTM-AE} GRU-AE AE SDAE ]]]};\\
};
\end{tikzpicture}
\caption{   {Autoencoder architectures for anomaly detection}.
        \\AE: Autoencoders~\cite{liou2014autoencoder}, LSTM : Long Short Term Memory Networks~\cite{hochreiter1997long}
        \\SDAE: Stacked Denoising Autoencoder~\cite{vincent2010stacked}, DAE : Denoising Autoencoders~\cite{vincent2010stacked}
        \\GRU: Gated Recurrent Unit~\cite{cho2014learning}, CNN: Convolutional Neural Networks~\cite{krizhevsky2012imagenet}
        \\CNN-LSTM-AE: Convolution Long Short Term Memory Autoencoders~\cite{haque2018image}
        \\CAE: Convolutional Autoencoders~\cite{masci2011stacked}
        }

 \label{fig:aevariants}
\end{figure}

\section{ Relative Strengths and Weakness : Deep Anomaly Detection Methods}
\label{sec:relativeSOW}
Each of the deep anomaly detection (DAD) techniques discussed in previous sections have their unique strengths and weaknesses. It is critical to understand which anomaly detection technique is best suited for a given anomaly detection problem context. Given the fact that DAD is an active research area, it is not feasible to provide such an understanding for every anomaly detection problem. Hence in this section, we analyze the relative strengths and weaknesses of different categories of techniques for a few simple problem settings. Classification based supervised DAD techniques illustrated in Section \ref{sec:supervisedDAD} are better choices in scenario consisting of the equal amount of labels for both normal and anomalous instances. The computational complexity of supervised DAD technique is a key aspect, especially when the technique is applied to a real domain. While classification based, supervised or semi-supervised techniques have expensive training times, testing is usually fast since it uses a pre-trained model. Unsupervised DAD techniques presented in Section~\ref{sec:unsupervisedDAD} are being widely used since label acquisition is a costly and time-consuming process. Most of the unsupervised deep anomaly detection requires priors to be assumed on the anomaly distribution hence the models are less robust in handling noisy data. Hybrid models illustrated in Section~\ref{sec:hybridModels} extract robust features within hidden layers of the deep neural network and feed to best performing classical anomaly detection algorithms. The hybrid model approach is suboptimal because it is unable to influence representational learning in the hidden layers. The One-class Neural Networks (OC-NN) described in Section~\ref{sec:oneclassNN} combines the ability of deep networks to extract a progressively rich representation of data along with the one-class objective, such as a hyperplane ~(\cite{chalapathy2018anomaly}) or hypersphere ~(\cite{ruff2018deep}) to separate all the normal data points from  anomalous data points. Further research and exploration is necessary to comprehend better the benefits of this new architecture proposed.

\section{Conclusion}
\label{sec:chapter1_conclusion}
In this survey paper, we have discussed various research methods in deep learning-based anomaly detection along with its application across various domains. This article discusses the challenges in deep anomaly detection and presents several existing solutions to these challenges. For each category of deep anomaly detection techniques, we present the assumption regarding the notion of normal and anomalous data along with its strength and weakness. The goal of this survey was to investigate and identify the various deep learning models for anomaly detection and evaluate its suitability for a given dataset. When choosing a deep learning model to a particular domain or data, these assumptions can be used as guidelines to assess the effectiveness of the technique in that domain. Deep learning based anomaly detection is still active research, and a possible future work would be to extend and update this survey as more sophisticated techniques are proposed.







\bibliographystyle{unsrtnat}
\bibliography{main}

\end{document}